%% file: sn-article.tex
\documentclass[sn-basic]{sn-jnl}

\usepackage{graphicx}%
\usepackage{multirow}%
\usepackage{amsmath,amssymb,amsfonts}%
\usepackage{amsthm}%
\usepackage{mathrsfs}%
\usepackage[title]{appendix}%
\usepackage{xcolor}%
\usepackage{textcomp}%
\usepackage{manyfoot}%
\usepackage{booktabs}%
\usepackage{algorithm}%
\usepackage{algorithmicx}%
\usepackage{algpseudocode}%
\usepackage{listings}%
\usepackage{subfigure}
\usepackage{tikz}
\usepackage{pgfplots}

\theoremstyle{thmstyleone}%
\newtheorem{theorem}{Theorem}
\theoremstyle{thmstyletwo}%

\theoremstyle{thmstylethree}%
\newtheorem{conjecture}[theorem]{Conjecture}
\def\ci{\perp\!\!\!\perp}
\begin{document}

\title[Dropout Drops Double Descent]{Dropout Drops Double Descent}


\author*[1]{\fnm{Tian-Le} \sur{Yang}}\email{yangtianle1996@gmail.com}

\author[1]{\fnm{Joe} \sur{Suzuki}}\email{prof.joe.suzuki@gmail.com}


\affil[1]{\orgdiv{Graduate School of Engineering Science}, \orgname{Osaka University}}





\abstract{
This study demonstrates that double descent can be mitigated by adding a dropout layer adjacent to the fully connected linear layer. The unexpected double-descent phenomenon garnered substantial attention in recent years, resulting in fluctuating prediction error rates as either sample size or model size increases. Our paper posits that the optimal test error, in terms of the dropout rate, shows a monotonic decrease in linear regression with increasing sample size. Although we do not provide a precise mathematical proof of this statement, we empirically validate through experiments that the test error decreases for each dropout rate. The statement we prove is that the expected test error for each dropout rate within a certain range decreases when the dropout rate is fixed. Our experimental results substantiate our claim, showing that dropout with an optimal dropout rate can yield a monotonic test error curve in nonlinear neural networks. These experiments were conducted using the Fashion-MNIST and CIFAR-10 datasets. These findings imply the potential benefit of incorporating dropout into risk curve scaling to address the peak phenomenon. To our knowledge, this study represents the first investigation into the relationship between dropout and double descent.
}


\maketitle

\section{Introduction}\label{sec1}

\label{intro}
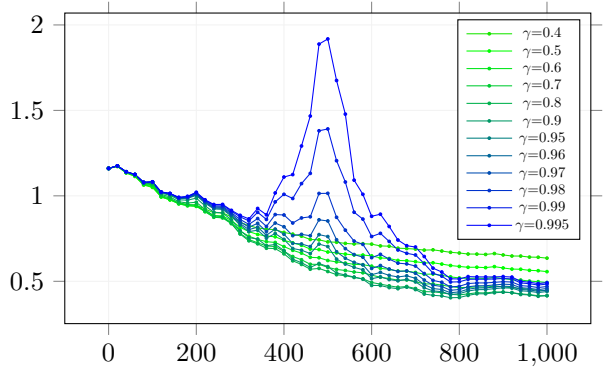
\begin{figure}[!ht]
\centering
\input{fig1}
\caption{\textbf{Test Risk of Sample-Wise Double Descent with Dropout.} $\gamma$ denotes the probability of dropout as $R$. The number in the legend is the present probability. $p\ =\ 500$ and the sample size of the x-axis. Here $x\sim\mathcal{N}(0,I_p)$, $\mathrm{y}=x^\top\beta^*+\epsilon$, $\epsilon\sim\mathcal{N}(0,0.25),\ \beta^*\sim\mathcal{U}(0,1)$ and $||\beta^*||_2=1$.}\label{fig:1}
\end{figure}

Recent investigations have shown that over-parameterized models, including linear regression and neural networks \citep{2,5,7,3,6,4,23}, demonstrate significant generalization capabilities, even when the labels are influenced by pure noise. This unique characteristic has attracted considerable academic attention, posing significant challenges to traditional generalization theory. A key framework, "Double Descent," helps explain this behavior \citep{2}. In the under-parameterized realm, as we increase the number of model parameters or sample sizes, the test error initially shows a reduction, as illustrated by the peak curve in Figure \ref{fig:1}. Intriguingly, as we transition into the over-parameterized domain, instead of increasing, the test error continues to decrease, revealing an unexpected secondary descent phase.

This peak phenomenon was first observed as early as three decades ago \citep{3,4}, and its re-emergence in recent years \citep{2,23} underlines the significant role it plays in research within the over-parameterized regime.

A primary objective of machine learning algorithms is to provide accurate out-of-sample predictions—a quality known as generalization. Traditional generalization theory presents a 'U-shaped' risk curve derived from the bias-variance trade-off \citep{1}, which suggests the optimal model selection occurs prior to the interpolation point (when $n=p$). This trade-off suggests that a small hypothesis class lacks the expressive power necessary to include the truth function. Conversely, a larger class may introduce spurious overfitting patterns. However, in contrast to this traditional view, the double-descent behavior, marked by a "$\verb|\/\|$"-shaped trend with increasing model size, implies that we can discover a superior model with zero train and test error without succumbing to overfitting.

The reason behind the relatively recent surge in attention towards the double descent phenomenon is somewhat elusive, but the widespread adoption of regularization methods, such as ridge regularization \citep{7,8} and early stopping \citep{9}, designed to nullify double descent, might provide some explanation. In this study, we focus on one of the most popular regularization methods—dropout.

Dropout is a well-established regularization technique for training deep neural networks. It aims to prevent 'co-adaptation' among neurons by randomly excluding them during training \citep{57}. Dropout's effectiveness extends across a wide range of machine learning tasks, from classification \citep{10} to regression \citep{58}. Notably, dropout was a vital component in the design of AlexNet \citep{56}, which significantly outperformed its competitors in the 2012 ImageNet challenge. Due to dropout's proven efficiency in avoiding overfitting \citep{10} and its broad application scope, we propose that it may significantly mitigate the double descent phenomenon. This leads us to the following question:

\begin{center}
	\textit{Under what conditions and how does dropout mitigate the double descent phenomenon?}
\end{center}

We recognize that the double-descent phenomenon exists under both sample-wise and model-wise conditions. This paper considers its occurrence in both linear and nonlinear models, to improve test performance without unexpected non-monotonic responses. The elimination of double descent has indeed become a hot research topic. For instance, ridge regularization can alleviate double descent \citep{8}, as can early stopping \citep{9}.

We explore a well-specified linear regression model utilizing dropout with ${r}_{ij}\sim\mathcal{B}er(\gamma),{r}\in\{0,1\}^{n\times p},\gamma>0,{X}\in\mathbb{R}^{n\times p}, y\in\mathbb{R}^n, {\beta}\in\mathbb{R}^{p}$, aiming to minimize the empirical risk:
$$	L=||{y}-({r}\ast {X}){\beta}||_2^2\ ,$$
where $\ast$ denotes an element-wise product, serving to drop parameters during the training phase randomly. Dropout aids in preventing overfitting and offers a means to efficiently combine a wide range of different neural network architectures \citep{10}.

\noindent\textbf{Our Contributions.} Our study tackles the aforementioned question using theoretical and empirical methodologies. Theoretically, we explore the simplest linear regression with dropout regularization, which echoes the influence observed in general ridge regression \citep{11}. When considering the test error—which includes both the bias and variance of a well-formulated linear regression model that employs dropout for isotropic Gaussian features\footnote{Normal distribution with an identity covariance matrix.}—we adopt a non-asymptotic perspective. Although we couldn't secure an exact solution to substantiate the monotonic decline of the test error, we devised an alternative approach. Through the application of Taylor series expansion, we obtained an approximate solution, providing persuasive evidence supporting the continuous decrease of the test error. On the empirical front, our numerical experiments demonstrate that the dropout technique can effectively mitigate the double descent phenomenon in both linear and nonlinear models. In more specific terms, we demonstrate:

\begin{itemize}
\item \textbf{Eliminating the Sample-Wise Double Descent.} We empirically validate the monotonicity of the test error as the sample size increases (see Figure \ref{fig:1}) and theoretically prove the monotonicity of the second-order Neumann series test error. We plan to detail the exact solution in future work.
\item \textbf{Eliminating the Model-Wise Double Descent.} We empirically demonstrate the monotonicity of the test error as the model size increases.
\item \textbf{Multi-layer CNN.} We provide empirical evidence showing that dropout can alleviate the double descent in multi-layer CNNs.
\end{itemize}

\subsection{Related works}
\label{sec:1}
\textbf{Dropout.} The purpose of dropout, as proposed in \cite{10}, is to alleviate overfitting, and numerous variants of this technique have been further examined in \cite{31,32,33,36,39,40,41}. As for the theory behind dropout, \cite{13} demonstrates that it functions as an adaptive regularization. \cite{14} postulates that dropout operates akin to a Bayesian approximation algorithm—specifically a \textit{Gaussian Process}, incorporating an element of uncertainty into the functioning of black-box neural networks. Additionally, several studies have addressed the Rademacher complexity of dropout \citep{15}, and its implicit and explicit regularization \citep{16,38}.

\noindent\textbf{Generalized Ridge Regression.} The dropout estimator resembles a generalized ridge estimator, represented as $\hat{{\beta}}=({X}^\top{X}+\lambda{\Sigma_w})^{-1}{X}^\top{y}$, with ${\Sigma_w}$ being the weighted matrix and $\lambda>0$. Generalized ridge regression was first introduced in \cite{25}, with numerous developments discussed in \cite{28,19,26,11,29,30,27}. Nevertheless, these estimators are typically contemplated when $n > p$. Hence, their impact in high-dimensional and over-parameterized regimes is scarcely known. \cite{20} recently provided an asymptotic view of the weighted $\ell_2$ regularization in linear regression.

\noindent\textbf{Dropping Double Descent.} Several studies have aimed to counteract the double descent phenomenon. \cite{9} illustrates that early stopping can attenuate double descent. \cite{8} argues that optimal ridge regularization has a similar effect in the non-asymptotic view, a finding that aligns with our study. \cite{7} further sheds light on ridge regularization, illustrating a trend towards the same test error as the tail of double descent in model size.
\section{Background}
\label{sec:2}
We consider linear regression in which $p$ ($\geq 1$) covariates $x\in {\mathbb R}^p$ and response $\mathrm{y}\in {\mathbb R}$ are related by
\begin{equation}\label{eqj-1}
\mathrm{y} = {x}^\top\beta_0+\epsilon\ ,\ \epsilon\sim\mathcal{N}(0,\sigma^2)
\end{equation}
with unknown ${\beta}_0\in\mathbb{R}^p$ and $\sigma^2>0$, where the occurrences of $\epsilon$ is independent from those of $x$, and we estimate $\beta_0$ from $n$($\geq 1$) i.i.d. training data $(x_1,\mathrm{y}_1),\ldots,(x_n,\mathrm{y}_n) \in {\mathbb R}^p\times {\mathbb R}$.

In particular, we assume that the covariates are generated by  
\begin{equation}\label{eqj-2}
{x} \sim \mathcal{N}(0,\ I_p)\ .
\end{equation}
 Thus, the covariates and response have the joint distribution $\mathcal D$ defined by (\ref{eqj-1}) and (\ref{eqj-2}), and
we express $z^n:=\{({x}_i,\mathrm{y}_i)\}^n_{i=1}\sim \mathcal{D}^n$ for the training data.
For each $\beta\in {\mathbb R}^p$, we define 
\begin{equation}\label{eqj-3}
	R({\beta}) := \mathop{\mathbb{E}}\limits_{({x,\mathrm{y}})\sim\mathcal{D}}[({x}^\top{\beta}-{\mathrm{y}})^2],
\end{equation}
where $\mathop{\mathbb{E}}\limits_{(x,\mathrm{y})\sim\mathcal{D}}[\cdot]$ is the expectation  w.r.t. the distribution $\mathcal{D}$.

Suppose we estimate $\beta$ from the training data $z^n$ by 
$\hat{\beta}_n: ({\mathbb R}^p\times {\mathbb R})^n\rightarrow {\mathbb R}^p$.
Then, we define 
\begin{equation}\label{eqj-4}
\begin{aligned}
    	\bar{R}(\hat{\beta}_n) :&= \mathop{\mathbb{E}}\limits_{z^n \sim\mathcal{D}^n}R(\hat{\beta}_n(z^n))
	= 
	\mathop{\mathbb{E}}\limits_{z^n \sim\mathcal{D}^n}\mathop{\mathbb{E}}\limits_{({x,\mathrm{y}})\sim\mathcal{D}}[({x}^\top{\hat{\beta}_n(z^n)}-{\mathrm{y}})^2]
\end{aligned}
\end{equation}
where 
$\displaystyle {\mathbb E}_{z^n\sim\mathcal{\cal D}^n}[\cdot]$ is
the expectation  w.r.t. the distribution $D^n$.
Note that (\ref{eqj-4}) averages (\ref{eqj-3}) over the training data as well while both evaluate the expected squared loss of the estimation.

In this paper, we consider the situation of dropout:
given the training data $z^n=\{(x_i,\mathrm{y}_i)\}_{i=1}^n$, 
for $X=[x_1,\ldots,x_n]^\top\in {\mathbb R}^{n\times p}$ and 
$y=[\mathrm{y}_1,\ldots,\mathrm{y}_n]^\top\in {\mathbb R}^n$,
we estimate $\beta$ by the $\hat{\beta}(z^n)$ that minimizes
the training error
$\mathop{\mathbb{E}}\limits_{r\sim\mathcal{B}er(\gamma)}[L]$ for
$$L=\|{y}-({r}\ast {X}){\beta}\|_2^2\ ,$$
where $\ast$ denotes the element-wise product, each element of $R\in {\mathbb R}^{n\times p}$ takes one and zero with probabilities $\gamma$ and $1-\gamma$, respectively, and we write $r\sim Ber(\gamma)$ for the distribution. Then, the quantity $\mathop{\mathbb{E}}\limits_{r\sim\mathcal{B}er(\gamma)}[L]$ can be expressed by 
\begin{equation}\label{eqj-12}
\begin{aligned}
&\mathop{\mathbb{E}}\limits_{r\sim\mathcal{B}er(\gamma)}\|{y}-(r* X){\beta}\|_2^2=\mathop{\mathbb{E}}\limits_{r\sim\mathcal{B}er(\gamma)}\|{y}-M{\beta}||_2^2\\
		&={y}^\top{y}-2{\beta}^\top\mathbb{E}(M^\top){y}+{\beta}^\top\mathbb{E}(M^\top M){\beta}\\
		&={y}^\top{y}-2\gamma{\beta}^\top X^\top{y}+{\beta}^\top\mathbb{E}(M^\top M){\beta}\\
		&=\|{y}-\gamma X{\beta}\|_2^2-\gamma^2{\beta}^\top X^\top X{\beta}+{\beta}^\top\mathbb{E}(M^\top M){\beta}\\
		&=\|{y}-\gamma X{\beta}\|_2^2+{\beta}^\top(\mathbb{E}(M^\top M)-\gamma^2X^\top X){\beta}\\
		&=\|{y}-\gamma X{\beta}\|_2^2+(1-\gamma)\gamma\|\Gamma{\beta}\|_2^2
	\end{aligned}
\end{equation}
where $M:=r\ast X$, $\Gamma=\mathrm{diag}(X^\top X)^{1/2}$,
the final equation follows from the fact that
the element-wise expectation ${\mathbb E}(M^\top M)$ is 
$${\mathbb E}\left[\sum_{k}m_{ik}m_{jk}\right]=
\left\{
\begin{array}{ll}
\gamma^2\sum_{k}x_{ik}x_{jk},&i\not=j\\
\vspace{1mm}
\gamma\sum_{k}x_{ik}^2,&i=j\\
\end{array}
\right.\ $$
for the $(i,j)$-th element of $M^\top M$ 
(the off-diagonal elements of ${\mathbb E}(M^\top M)$ and $\gamma^2X^\top X$ are canceled out).

 We can consider this as a Tikhonov regularization method. Let $\beta'=\gamma\beta$ as in \cite{10}. Then, (\ref{eqj-12}) becomes 
\begin{eqnarray}\label{eq55}
\|y-X\beta'\|^2+\frac{1-\gamma}{\gamma}\|\Gamma\beta'\|^2\ ,
\end{eqnarray}
which is minimized when $\beta'$ is equal to 
\begin{equation}\label{eq56}
\hat{{\beta}}_{n,\gamma}=\left(X^\top X+\frac{1-\gamma}{\gamma}\Gamma^\top \Gamma\right)^{-1}X^\top{y}\ .
\end{equation}

\section{Drop Double-Descent in Linear Regression}

In this section, we show the monotonicity of the solution in the sample size $n$ with dropout in linear regression, and its proof follows in Appendix~\ref{approx}.
Hereafter, we denote $\hat{\beta}$ by $\hat{\beta}_{n,\gamma}$ when we require $n$ and $\gamma$ to be explicit.

Before proving the claim, we notice that the test error is of the form
\begin{eqnarray}
&&{R}(\hat{\beta})=
\mathop{\mathbb{E}}\limits_{({x},y)\sim\mathcal{D}}\left[\{{x}^\top(\hat{{\beta}}-\beta_0)+\epsilon\}^2\right]\nonumber
=
\|\hat{{\beta}}-{\beta}_0\|_2^2+\sigma^2\label{eqj-14}\ ,
\end{eqnarray}
which is due to
\begin{eqnarray*}
\mathop{\mathbb{E}}\limits_{{x}\sim\mathcal{N}(0,I_d),\epsilon\sim\mathcal{N}(0,\sigma^2)}
[\{ (\hat{{\beta}}-{\beta}_0)^\top x+\epsilon\}^2]
=
\mathop{\mathbb{E}}\limits_{{x}\sim\mathcal{N}(0,I_p)}
[(\{(\hat{\beta}-{\beta}_0)^\top x\})^\top\{ (\hat{\beta}-{\beta}_0)^\top x\}]
+\sigma^2 .
\end{eqnarray*}
For the dropout estimator Eq.~(\ref{eq56}), the expected test error is
\begin{eqnarray}\label{eqj-20}
\bar{R}(\hat{\beta}_{n,\gamma})\nonumber
	&=&\mathbb{E}_X\mathbb{E}_y[{R}(\hat{\beta}_{n,\gamma})]={\mathbb{E}}_X\mathbb{E}_{y}[\|\hat{\beta}_{n,\gamma}-\beta_0\|_2^2]+\sigma^2\nonumber\\	&=&
 \mathbb{E}_{X}\mathbb{E}_{y}[\|( X^\top X+\Lambda)^{-1}X^\top y-\beta_0\|_2^2]+\sigma^2\nonumber\\
		&=&\mathbb{E}_{X}[\|( X^\top X+\Lambda)^{-1}X^\top(X\beta_0+\epsilon)-\beta_0\|_2^2]+\sigma^2\nonumber\\
		&=&\mathbb{E}_{X}[\|((X^\top X+\Lambda)^{-1}X^\top X-I_p)\beta_0\|_2^2]\nonumber+\sigma^2{\mathbb{E}}_{X}[\|(X^\top X+\Lambda)^{-1}X^\top\|_F^2]+\sigma^2\nonumber
\end{eqnarray}
where $\Lambda=\frac{1-\gamma}{\gamma}\mathrm{diag}(X^\top X)$. By neglecting the constant terms, the quantity $\bar{R}(\hat{\beta}_{n,\gamma})$ becomes
\begin{equation}\label{eq1}
\begin{aligned}
     &\beta_0^{\top}\mathbb{E}_{X}\left[\left(I+A^\top\right)^{-1}\left(I+A\right)^{-1}\right]\beta_0+\sigma^{2} {\mathbb{E}}_{X}\left[\left\|\left(X^{\top} X+\Lambda\right)^{-1} X^{\top}\right\|_{F}^{2}\right],
\end{aligned}
\end{equation}
where $A=\Lambda^{-1}X^\top X$.


We evaluate the expected test error (\ref{eq1}) by taking Taylor's expansion of the matrix 
$$\left(I+A^\top\right)^{-1}\left(I+A\right)^{-1}\ .$$
Then, we claim\footnote{We say $f(n)=O(g(n))$ if there exist $b>0$ and $n_0\geq 1$ such that $|f(n)|\leq b|g(n)|$ for $n \geq n_0$.}. 
\begin{theorem}\label{thm1}
Let $\alpha=C<\frac{1}{(1+\sqrt{\frac{p}{n}})^2}$, the expected test error (\ref{eq1}) is
\begin{equation*}
\begin{aligned}
f(\alpha)=&\left\{1-2\alpha+3\alpha^{2} \frac{p}{n}\right\}\left\|\beta_0\right\|^{2}+\sigma^2\alpha^2(\alpha+1)\frac{p}{n}+O({\frac{1}{n^2}})
\end{aligned}
\end{equation*}
with ${\alpha=\frac{\gamma}{1-\gamma}}$.
\end{theorem}
To make the expected test error monotonically decrease with the chosen hyperparameter $\alpha$, we need to consider the expected test error in Theorem~\ref{thm1} will not be divergent. To ensure the convergence of this Neumann series, we need the eigenvalue of $\alpha\cdot A=\alpha\cdot\Lambda^{-1}X^\top X$ to be smaller than $1$. Therefore, we need to consider the largest eigenvalue $\lambda_{\text{max}}$ of $A$ to make $\alpha < \frac{1}{\lambda_{\text{max}}}$. Before this, we notice some critical points. 
\begin{enumerate}
\item 
Let $Q:=\mathrm{diag}(X^\top X)$, 
$P:=Q^{-\frac{1}{2}}X^\top XQ^{-\frac{1}{2}}$, 
$\Lambda:=\frac{1-\gamma}{\gamma}Q$, and 
$M:=\Lambda^{-\frac{1}{2}}X^\top X\Lambda^{-\frac{1}{2}}$.
Then, $M$ and $A=\Lambda^{-1}X^\top X$ share share the same characteristic polynomial
\begin{equation*}
    \begin{aligned}
        &\mathcal{P}_{M}(\lambda)
        =\mathrm{det}(\Lambda^{-\frac{1}{2}}X^\top X\Lambda^{-\frac{1}{2}}-\lambda I)
        =
        \mathrm{det}(\Lambda^{-1/2})\mathrm{det}(X^\top X-\Lambda^{\frac{1}{2}}\lambda \Lambda^{\frac{1}{2}})\mathrm{det}(\Lambda^{-\frac{1}{2}})\\
        &=\mathrm{det}(\Lambda^{-1})
        \mathrm{det}(X^\top X-\lambda \Lambda)=\mathrm{det}(\Lambda^{-1}X^\top X-\lambda I)=\mathcal{P}_{A}(\lambda)
    \end{aligned}\ ,
\end{equation*}
so do the eigenvalues.

\item 
Let $\lambda_{\mathrm{max}}$ and $\lambda_{\mathrm{min}}$
be the largest and smallest eigenvalues of $M$. Then, 
    $\lambda_{\mathrm{max}}\rightarrow(1+\sqrt{d})^2$ and $\lambda_{\mathrm{min}}\rightarrow(1-\sqrt{d})^2$ 
    as $n,p\rightarrow\infty $ with $\frac{p}{n}\rightarrow d\in(0,\infty)$ if $\mathbb{E}[x^4]<\infty$ (Theorem 1.1 in \cite{45}).
\end{enumerate}

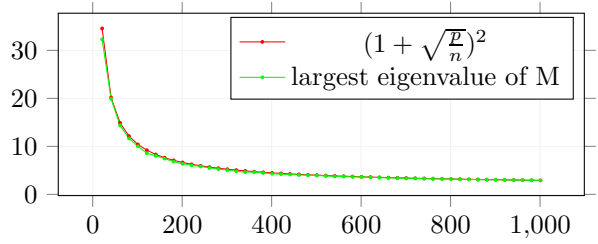
\begin{figure}[!ht]
\centering
\input{fig5}
\caption{\textbf{The Largest eigenvalue of Sample Correlation Matrix ($Q\in\mathbb{R}^{n\times p}$)}. X-axis denotes the number of sample $n$, Y-axis denotes the magnitude of largest eigenvalue and $n\in\mathbb{N},\ p=500 $}\label{fig:5}
\end{figure}

Hence, the maximum eigenvalues of matrices $M$ and $A$ are shown to approach $(1+\sqrt{\frac{p}{n}})^2$ asymptotically. 
Moreover, our empirical investigations corroborate that the largest eigenvalue of the sample correlation matrix $M$ aligns closely with the theoretical prediction of $(1+\sqrt{\frac{p}{n}})^2$, as illustrated in Fig.~\ref{fig:5}. The Taylor series expansion converges when the parameter $\gamma/(1-\gamma)$ is multiplied to make the largest eigenvalue of M less than 1. The proof of Theorem~\ref{thm1} is in Appendix~\ref{approx}.  
\section{Experiments}
{
This section provides empirical evidence that dropout with optimal rate can effectively eliminate the double descent phenomenon in a broader range of scenarios compared to what is formally proven in Theorem~\ref{thm1}.}

\subsection{Monotonicity for Sample-wise Double Descent}
\textbf{Elimination Double Descent in Linear Regression. (Synthetic Data)}\\
In this part, we evaluate test error using dropout with pseudo optimal probability 0.8 (from Figure~\ref{fig:1}) in linear regression, the sample distribution $x\sim\mathcal{N}(0,I_p)$, $y=x^\top \beta^*+\epsilon$, $\epsilon\sim\mathcal{N}(0,0.25),\ \beta^*\sim\mathcal{U}(0,1)$ and $\|\beta^*\|_2=1$. Moreover, the monotonic curves in Figure \ref{fig:2} show that the test error always remains monotonicity within the optimal dropout rate when the sample size increases for various dimensions $p$. 

\noindent\textbf{Random ReLU Initialization. (Fashion-MNIST)} \\
 We consider the random nonlinear features stemming from the random feature framework of \cite{17}. We apply random features to Fashion-MNIST \citep{18}, an image classification dataset with 10 classes. In the preprocessing step, the input images vector $x\in\mathbb{R}^d$ are normalized and flattened to $[-1,1]^d$ for the $d=784$. To make the correct estimation of mean square loss, the class labels are dealt with the one-hot encoding to $y\in\{\vec{e}_1,\dots,\vec{e}_{10}\}\subset\mathbb{R}^{10}$. According to the given number of random features $D$, and the number of sample data $n$, we are going to acquire the random classifier by performing linear regression on the nonlinear embedding:
	$\tilde{X}:=\mathrm{ReLU}(XW^\top )$
where $X\in\mathbb{R}^{n\times d}$ and $W\in\mathbb{R}^{D\times d}$ is a matrix with every entry sampled i.i.d from $\mathcal{N}(0,1/\sqrt{d})$, and with the nonlinear activation function ReLU applied pointwise. This is equivalent to a 2-layer fully connected neural network with a
frozen (randomly initialized) first layer, trained with dropout. Figure \ref{fig:3} shows the monotonic test error.
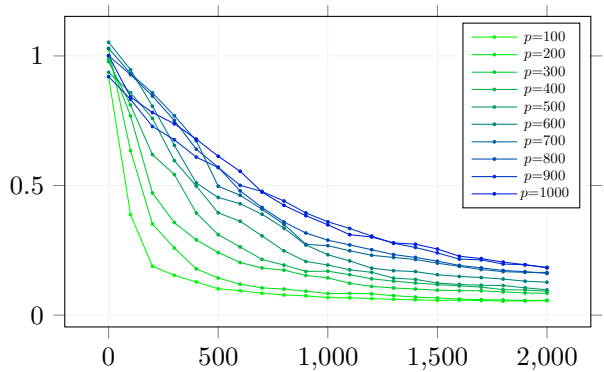
\begin{figure}[!ht]
\centering
\input{fig2}
\caption{\textbf{Test Risk with Number of Sample in linear regression with Dropout probability 0.8.} The test error curves decrease with the optimal dropout rate. The X-axis in this figure is the dimension of the parameter (0.8 is a pseudo-optimal value). The Y-axis is test risk.}\label{fig:2}
\end{figure} 
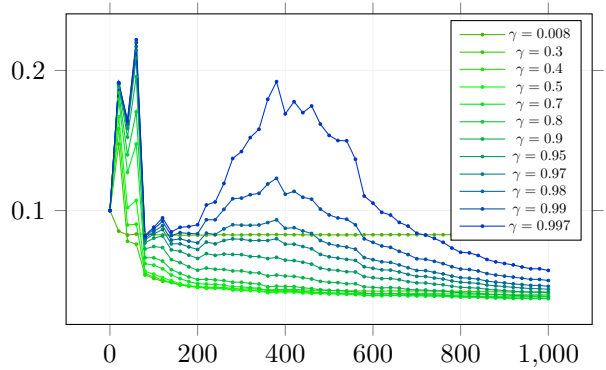
\begin{figure}[!ht]
\centering
\input{fig3}
\caption{\textbf{Test Risk with Number of Sample in Nonlinear Model with Dropout using Fashion-Mnist.} The test error curves are decreasing with the optimal dropout rate. X-axis: sample size; Y-axis: Test risk.}\label{fig:3}
\end{figure}

\subsection{Monotonicity for Model-wise Double Descent}
\begin{figure}[!ht]
\centering
\input{fig41}
\caption{\textbf{Test Risk with of model size in Linear Regression with Dropout.} The test error curves decrease with the optimal dropout rate. X-axis: the dimension of the parameter; Y-axis: Test risk.}\label{fig:4}
\end{figure}
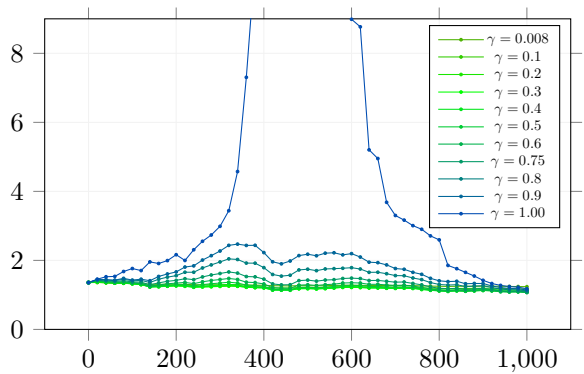

Like above setting, the sample distribution $x\sim\mathcal{N}(0,I_p)$, $y=x^\top \beta^*+\epsilon$, $\epsilon\sim\mathcal{N}(0,0.25),\ \beta^*\sim\mathcal{U}(0,1)$, $\|\beta^*\|_2=1$ and we fix $n=500$. The experiment result is the monotonic curves in Figure \ref{fig:4} show that the test error remains monotonicity with the optimal dropout rate as the model size increases. For the multiple descents in Figure \ref{fig:4}, the readers can find more details in \cite{55}. 

\subsection{Multi-layer CNN}
We use the same setups as in \cite{6}. Here, we give the brief details of the model. For the full details, please check Appendix \ref{model2}. 

\textbf{Standard CNNs:} We consider a simple family of 5-layer CNNs, with $4$ convolutional layers of  widths [$k,2k,4k,8k$] for varying $k$,  and  a  fully-connected  layer. For context, the CNN with width $k= 64$, can reach over $90\%$ test accuracy on CIFAR-10 with data augmentation. We train with cross-entropy loss and the following optimizer: Adam with $0.0001$ learning rate for 10K epochs; SGD with $0.1 / \sqrt{\lfloor T / 512\rfloor+1}$ for 500K gradient steps. 

\textbf{Label Noise.} In our experiments, label noise \citep{54} of probability prefers to train on samples with the correct label with probability $0\%,20\%$, and a uniformly random incorrect label otherwise (label noise is sampled only once and not per epoch).

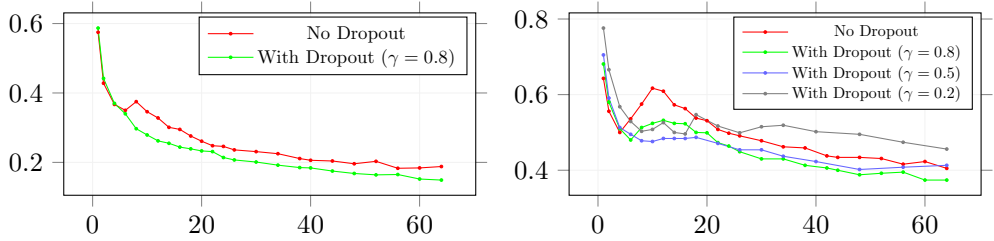
\begin{figure}[!ht]
\centering
\input{fig6}
\input{fig7noise20.tex}
\caption{\textbf{Test Risk with Number of width parameter in 5 layer-CNN with Dropout.} The x-axis is CNN width parameter (left: $0\%$ label noise with Adam; right: $20\%$ label noise with SGD). We can see dropout drops double descent.($\gamma$: present rate)}\label{fig:sgd20}
\end{figure}

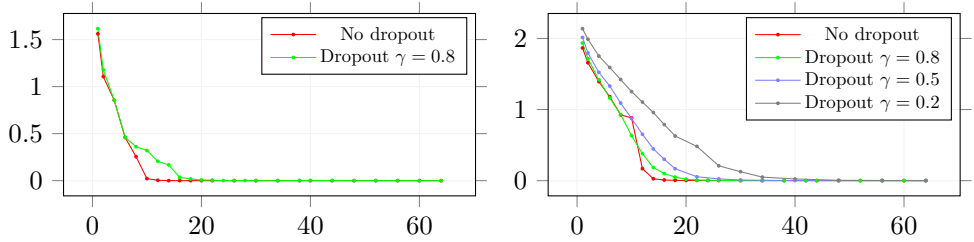
\begin{figure}[!ht]
\centering
\input{fig7traincnn}
\input{fig8train20}
\caption{\textbf{Train Loss with width parameter in 5 layer-CNN with Dropout (left: Adam, right: SGD). X-axis is CNN width parameter}}\label{fig:trainsgd20}
\end{figure}

\textbf{Dropout layer.} We add the dropout layer before the full-connected linear layer with the present rate $\gamma$ \citep{10}. Figure \ref{fig:sgd20} shows the test error results. The training loss is in Figure \ref{fig:trainsgd20}.

We observe model-wise double descent most strongly in settings with label noise in the train set (as is often true when collecting train data in the real world).
For model sizes at the interpolation threshold, there is effectively
only one model that fits the train data, and this interpolating model is very sensitive to noise in the
train set and/or model misspecification. That is, since the model can barely fit the train
data, forcing it to fit even slightly noisy or misspecified labels will destroy its global structure and
result in high test error. (See Figure 28 in the Appendix of \cite{6} for an experiment demonstrating this noise
sensitivity by showing that ensembling helps significantly in the critically parameterized regime). However, for over-parameterized models, many interpolating models fit the train set, and SGD can find one that “memorizes” (or “absorbs”) the noise while still performing well
on the distribution.

\section{Discussion} 
Our proof considers only the non-exact solution for expected test error, 
and therefore we cannot definitively assert that the test risk decreases monotonically. 
However, based on our experimental results and this non-exact proof, we propose the following conjecture:

\begin{conjecture}\label{tm2}
	\textit{
For any $n,p\geq 1$, $\sigma^2>0$, and $\beta_0$, the expected test risk is monotonic in sample as}
	\begin{equation}\label{eqj-10}
		\bar{R}(\hat{{\beta}}_{n+1})\leq\bar{R}(\hat{{\beta}}_{n}).
	\end{equation}
\end{conjecture}

In future research, we aim to prove that the exact solution with dropout can mitigate double descent.

Note that the optimal hyperparameter remains in the fixed dimension $p$ with a changeable sample size $n$. This is because the original data $y$ from the model $y=X\beta+\epsilon$ will change, thus affecting the common test error. Additionally, \cite{22} contains a statement about the sample covariance matrix $\mathrm{diag}(X^\top X)$, which converges to the identity matrix for all $\delta>0$ and $||x_i||_2\leq \sqrt{d}$ (Corollary 6.20 in \cite{22}):
\begin{equation}
	P[\|\frac{\mathrm{diag}(X^\top X)}{n}-I_p\|_2\geq\delta]\leq2p\cdot exp\left(-\frac{n\delta^2}{2d(1+\delta)}\right)
\end{equation}
for the $\mathbb{E}(\mathrm{diag}(X^\top X/n))=I_p$, and by coupling the previous conclusions, it seems that the dropout estimator tends to the ridge estimator \citep{12} and has the same asymptotic risk as the ridge estimator in \cite{7}.



Just as with dropout, the implementation of batch normalization \citep{59} is uncomplicated—it merely requires the incorporation of batch normalization layers into the network architecture. Its inherent simplicity positions batch normalization as an ideal candidate for expediting the training process associated with varying combinations of hyperparameters required to optimize the use of dropout layers. While this may not necessarily accelerate each training epoch, it's likely to facilitate swifter convergence. Given their similarities, several research studies have compared the two techniques \citep{60,61,62}. Based on this, we posit that Batch Normalization might also hold the capacity to alleviate the double descent phenomenon.

\section{Conclusion}
    Our study employs theoretical and empirical methods to investigate the impact of dropout regularization in linear regression. Theoretically, we extend our analysis to general ridge regression, adopting a non-asymptotic approach to understand the behavior of test error in linear regression models with dropout for isotropic Gaussian features. Empirically, we demonstrate through numerical experiments that dropout effectively mitigates the double descent phenomenon in linear and nonlinear models, including multi-layer CNNs. Our key contributions include demonstrating the elimination of sample-wise and model-wise double descent and providing evidence of dropout efficacy in multi-layer CNNs. For our future work, we will not only pay attention to the exact solution of the expected test risk but also consider the nonisotropic linear regression, even the theoretical analysis for multi-layer neural networks.

\backmatter

\begin{appendices}

\section{Proof}\label{secA1}
\subsection{Proof of Theorem~\ref{thm1}}\label{approx}

\noindent\textbf{The First term of (\ref{eq1})}\\
Let  $\Lambda:=\frac{1-\gamma}{\gamma} \operatorname{diag}\left(X^\top X\right)$, 
$A:=\Lambda^{-1} X^\top X$, 
and $\alpha = \frac{1-\gamma}{\gamma}$. 
We evaluate $\mathbb{E}[(I+ \left.A^{\top}\right)^{-1}(I+A)^{-1} ]$.
Note
	$	\left(I+A^{\top}\right)^{-1}(I+A)^{-1} 
		=I-A-A^{\top}+A^{2}+\left(A^{\top}\right)^{2}+A^{\top} A+\cdots$.
For $A=\left(a_{i, j}\right)$,  we have
$a_{i, j}=\frac{\gamma}{1-\gamma} \cdot \frac{\sum_{k} x_{k, i} x_{k, j}}{\sum_{k} x_{k, i}^{2}}$,
and  $\mathbb{E}[A]=\frac{\gamma}{1-\gamma}
\cdot I$, which is due to (\ref{eqj-2}). For  $A^{2}=\left(b_{i, j}\right)$,  
we have
\begin{equation*}
b_{i, j}=\left(\frac{\gamma}{1-\gamma}\right)^{2} \cdot \sum_{h} \frac{\sum_{k} x_{k, i} x_{k, h}}{\sum_{k} x_{k, i}^{2}} \frac{\sum_{k} x_{k, h} x_{k, j}}{\sum_{k} x_{k, h}^{2}}
\end{equation*}
and 
$\mathbb{E}\left[A^{2}\right]=\left(\frac{\gamma}{1-\gamma}\right)^{2} \frac{p}{n} \cdot I$. Apparently,
we have  $\mathbb{E}\left[A^{\top}\right]=\frac{\gamma}{1-\gamma}
\cdot I$  and  $\mathbb{E}\left[\left(A^{\top}\right)^{2}\right]=\left(\frac{\gamma}{1-\gamma}\right)^{2} \frac{p}{n} \cdot I$ ($\mathbb{E}[b_{i,j}]=0$ if $i\neq j$; if $i=j$, $\mathbb{E}[\hat{\rho}^2]=\frac{1}{n}$ if $\rho=0$).
Finally, we evaluate  $\mathbb{E}\left[A^{\top} A\right] $. For $A^{\top} A=\left(c_{i, j}\right)$, we have
\begin{equation*}
c_{i, j}=\left(\frac{\gamma}{1-\gamma}\right)^{2} \cdot \sum_{h} \frac{\sum_{k} x_{k, i} x_{k, h}}{\sum_{k} x_{k, h}^{2}} \frac{\sum_{k} x_{k, h} x_{k, j}}{\sum_{k} x_{k, h}^{2}}
\end{equation*}
so that $E\left[c_{i, j}\right]=0$ for $i \neq j$.
\begin{equation*}
\begin{aligned}
    \mathbb{E}\left[c_{i, i}\right]&=\left(\frac{\gamma}{1-\gamma}\right)^{2} \sum_{h}{\mathbb{E}}\left(\frac{\sum_{k} x_{k, i} x_{k, h}}{\sum_{k} x_{k, h}^{2}}\right)^{2}\\
    &={\left(\frac{\gamma}{1-\gamma}\right)^{2} \sum_{h}{\mathbb{E}}\left(\sum_{k} x_{k, i}\frac{x_{k, h}}{\sum_{k} x_{k, h}^{2}}\right)^{2}\quad(x_i\ci x_h)}\\
    &={\left(\frac{\gamma}{1-\gamma}\right)^{2} \sum_{h}{\mathbb{E}}\left(\frac{\sum_{k}x_{k, h}^2+2\sum_{i\neq j}x_{i,h}x_{j,h}}{(\sum_{k} x_{k, h}^{2})^2}\right)\quad(\mathbb{E}\left(2\sum_{i\neq j}x_{i,h}x_{j,h}\right)=0)}\\&=\left(\frac{\gamma}{1-\gamma}\right)^{2} \sum_{h} \mathbb{E}\left[\frac{1}{\sum_{k} x_{k, h}^{2}}\right]\ ,
\end{aligned}
\end{equation*}
where we have used
$\mathbb{E}\left(\sum_{r} u_{r} \alpha_{r}\right)^{2}={\mathbb{E} \sum_{r} u_{r}^{2} \alpha_{r}^{2}=\sum_{r} \alpha_r^{2}}$,
when  $u_{r} \sim N(0,1)$, $r=1,2, \cdots$,  are independent. Then, from the inverse density function of chi-square distribution,
we have
   $ \mathbb{E}\left[A^{\top} A\right] =\left(\frac{\gamma}{1-\gamma}\right)^{2} \cdot \frac{p}{n-2} \cdot I $.
Then, the first term of (\ref{eq1}) is
\begin{equation*}
\begin{aligned}
	\left\{1-2\left(\frac{\gamma}{1-\gamma}\right)\frac{p}{n}+\left(\frac{\gamma}{1-\gamma}\right)^{2}\left( \frac{2p}{n}+\frac{p}{n-2}\right)\right\}\left\|\beta_{0}\right\|^{2}\ .
\end{aligned}
\end{equation*}

\noindent\textbf{The Second term of (\ref{eq1})}\\
Since
\begin{equation*}
	\left\|\left(X^{\top} X+\Lambda\right)^{-1} X^{\top}\right\|_{F}^{2}=\operatorname{trace}\left\{\left(X^{\top} X+\Lambda\right)^{-1} X^{\top}\right\}^{\top}\left\{\left(X^{\top} X+\Lambda\right)^{-1} X^{\top}\right\}
\end{equation*}
the diagonal entries of
	$ X \Lambda^{-1}\left\{I-A^{\top}-A+A^{2}+\left(A^{\top}\right)^{2}+A^{\top} A\right\} \Lambda^{-1} X^{\top}$
are
\begin{equation*}
\begin{aligned}
	(X\Lambda^{-1}\Lambda^{-1}X^\top)\dots
	m_r^{\prime}&=\left(\frac{\gamma}{1-\gamma}\right)^{2}  \sum_{i}\frac{x_{r, i}^2}{(\sum_{k} x_{k, i}^{2})^2}\\
	(X\Lambda^{-1}A\Lambda^{-1}X^\top)\dots a_{r}^{\prime}&=\left(\frac{\gamma}{1-\gamma}\right)^{3} \sum_{i} \sum_{j} \frac{x_{r, i} x_{r, j} a_{i, j}}{\sum_{k} x_{k, i}^{2} \sum_{k} x_{k, j}^{2}} \\
	(X\Lambda^{-1}A^2\Lambda^{-1}X^\top)\dots b_{r}^{\prime}&=\left(\frac{\gamma}{1-\gamma}\right)^{4} \sum_{i} \sum_{j} \frac{x_{r, i} x_{r, j} b_{i, j}}{\sum_{k} x_{k, i}^{2} \sum_{k} x_{k, j}^{2}} \\
	(X\Lambda^{-1}A^\top A\Lambda^{-1}X^\top)\dots 
	c_{r}^{\prime}&=\left(\frac{\gamma}{1-\gamma}\right)^{4} \sum_{i} \sum_{j} \frac{x_{r, i} x_{r, j} c_{i, j}}{\sum_{k} x_{k, i}^{2} \sum_{k} x_{k, j}^{2}}\\
\end{aligned}
\end{equation*}
for $r=1, \ldots, n$.
First, we derive
\begin{equation*}
	\begin{aligned}
		\sum_{r}m_r^{\prime}&=\left(\frac{\gamma}{1-\gamma}\right)^{2}{\mathbb{E}} \sum_{i}\frac{\sum_{r}x_{r, i}^2}{(\sum_{k} x_{k, i}^{2})^2}=\left(\frac{\gamma}{1-\gamma}\right)^{2} \frac{p}{n-2}
	\end{aligned}
\end{equation*}
\begin{equation*}
\begin{aligned}
	\sum_{r} a_{r}^{\prime} &=\left(\frac{\gamma}{1-\gamma}\right)^3 \sum_{r} \sum_{i}\left\{\sum_{j \neq i} \frac{x_{r, i} x_{r, j} \frac{\sum_{k} x_{k, i} x_{k, j}}{\sum_{k} x_{k, i}^{2}}}{\sum_{k} x_{k, i}^{2} \sum_{k} x_{k, j}^{2}}+\frac{x_{r, i}^{2}}{\left(\sum_{k} x_{k, i}^{2}\right)^{2}}\right\} \\
	&=\left(\frac{\gamma}{1-\gamma}\right)^3 \sum_{i}\left\{\sum_{j \neq i} \frac{\left(\sum_{k} x_{k, i} x_{k, j}\right)^{2}}{\left(\sum_{k} x_{k, i}^{2}\right)^{2} \sum_{k} x_{k, j}^{2}}+\frac{1}{\sum_{k} x_{k, i}^{2}}\right\} \\
	&=\left(\frac{\gamma}{1-\gamma}\right)^3 \sum_{i} \frac{1}{\sum_{k} x_{k, i}^{2}}\left(\sum_{j \neq i} \hat{\rho}_{i, j}^{2}+1\right)
\end{aligned}
\end{equation*}
Please note that the distribution of $\hat{\rho}{i, j}$ is independent of $x{1, i}, \ldots, x_{n, i}$ (as demonstrated in the derivation).Hence, the expectation of $\sum_{r} a_{r}^{\prime}$ is
$\left(\frac{\gamma}{1-\gamma}\right)^3\left(\frac{p-1}{n}+1\right) \sum_{i} \frac{1}{\sum_{k} x_{k, i}^{2}}$, 
when $x_{1, i}, \ldots$ $x_{n, i}$ are given. Thus, we obtain
\begin{equation*}
E\left[\sum_{r} a_{r}^{\prime}\right]=\left(\frac{\gamma}{1-\gamma}\right)^3 \cdot \frac{p}{n-2} \cdot\left(\frac{p-1}{n}+1\right)
\end{equation*}
On the other hand.
\begin{equation*}
\sum_{r} b_{r}^{\prime}=\left(\frac{\gamma}{1-\gamma}\right)^4\sum_{r} \sum_{i} \sum_{j} \frac{x_{r, i} x_{r, j}}{\sum_{k} x_{k, i}^{2} \sum_{k} x_{k, j}^{2}} \sum_{h} \frac{\sum_{k} x_{k, i} x_{k, h}}{\sum_{k} x_{k, i}^{2}} \frac{\sum_{k} x_{k, h} x_{k, j}}{\sum_{k} x_{k, h}^{2}}
\end{equation*}
Let
\begin{equation*}
\beta_{i, j, h}:=\sum_{r} \frac{x_{r, i} x_{r, j}}{\sum_{k} x_{k, i}^{2} \sum_{k} x_{k, j}^{2}} \frac{\sum_{k} x_{k, i} x_{k, h}}{\sum_{k} x_{k, i}^{2}} \frac{\sum_{k} x_{k, h} x_{k, j}}{\sum_{k} x_{k, h}^{2}}
\end{equation*}
Then, the  $\sum_{h} \beta_{i, j, h}$  with  $i=j$  is
$\frac{1}{\sum_{k} x_{k, i}^{2}} \sum_{h}\left(\frac{\sum_{k} x_{k, h} x_{k, j}}{\sqrt{\sum_{k} x_{k, h}^{2} \sum_{k} x_{k, i}^{2}}}\right)^{2}$
and its expectation is  $\frac{1}{n-2}\left(\frac{p-1}{n}+1\right)$. When $j \neq i=h$, it's
$\frac{1}{\sum_{k} x_{k, i}^{2}}\left(\frac{\sum_{k} x_{k, i} x_{k, j}}{\sqrt{\sum_{k} x_{k, i}^{2} \sum_{k} x_{k, j}^{2}}}\right)^{2}$, 
its expectation is  $\frac{1}{n(n-2)}$ .  Since the  $\beta_{i, j, h}$  with  $i, j, h$  different is
\begin{equation*}
\sum_{r} \frac{x_{r, i} x_{r, j}}{\sum_{k} x_{k, i}^{2} \sum_{k} x_{k, j}^{2}} \frac{\sum_{k} x_{k, i} x_{k, h}}{\sum_{k} x_{k, i}^{2}} \frac{\sum_{k} x_{k, h} x_{k, j}}{\sum_{k} x_{k, h}^{2}}
\end{equation*}
its expectation is  $\frac{1}{n(n-2)}$  If we take expectation w.r.t.  $\left\{x_{k, h}\right\}$,  then the value becomes
\begin{equation*}
\sum_{r} \frac{x_{r, i} x_{r, j}}{\left(\sum_{k} x_{k, i}^{2}\right)^{2} \sum_{k} x_{k, j}^{2}} \cdot \frac{1}{n} \sum_{k} x_{k, i} x_{k, j}\ ,
\end{equation*}
where the fact  $E\left[\frac{Z_{1}}{Z_{1}+\cdots+Z_{m}}\right]=\frac{1}{m}$  for i.i.d.  $Z_{1}, \ldots, Z_{m}$  has been used. Thus, the expectation is  $\frac{1}{n(n-2)}$  as well. Hence,  $E\left[\sum_{h} \beta_{i, j, h}\right]$  with  $i \neq j$  is  $\frac{p}{n(n-2)}$ .  Therefore,
\begin{equation*}
E\left[\sum_{r} b_{r}^{\prime}\right]=\left(\frac{\gamma}{1-\gamma}\right)^4\frac{1}{n-2}\left(\frac{2p-1}{n}+1\right)\ .
\end{equation*}
Finally, we obtain  $E\left[\sum_{r} c_{r}^{\prime}\right]$ .  Let
\begin{equation*}
\gamma_{i, j, h}:=\sum_{r} \frac{x_{r, i} x_{r, j}}{\sum_{k} x_{k, i}^{2} \sum_{k} x_{k, j}^{2}} \frac{\sum_{k} x_{k, i} x_{k, h}}{\sum_{k} x_{k, h}^{2}} \frac{\sum_{k} x_{k, h} x_{k, j}}{\sum_{k} x_{k, h}^{2}} .
\end{equation*}
If  $i=j$,  we have
$\sum_{h} \gamma_{i, j, h}:=\sum_{h} \frac{1}{\sum_{k} x_{k, h}^{2}} \cdot\left(\frac{\sum_{k} x_{k, i} x_{k, h}}{\sqrt{\sum_{k} x_{k, i}^{2}} \sqrt{\sum_{k} x_{k, h}^{2}}}\right)^{2}$
and its expectation is  $\frac{d}{n(n-2)}$.  If  $i \neq j, h=i$
\begin{equation*}
\gamma_{i, j, h}:=\sum_{r} \frac{x_{r, i} x_{r, j}}{\sum_{k} x_{k, i}^{2} \sum_{k} x_{k, j}^{2}} \frac{\sum_{k} x_{k, i} x_{k, j}}{\sum_{k} x_{k, i}^{2}}=\frac{1}{\sum_{k} x_{k, i}^{2}}\left(\frac{\sum_{k} x_{k, i} x_{k, j}}{\sqrt{\sum_{k} x_{k, i}^{2}} \sqrt{\sum_{k} x_{k, j}^{2}}}\right)^{2}
\end{equation*}
and its expectation is  $\frac{1}{n(n-2)}$  If  $i, j, h$  are different, if we fix  $\left\{x_{k, j}\right\}$  and  $\left\{x_{k, h}\right\}$,  then the expectation of
$\gamma_{i, j, h}:=\frac{1}{\sum_{r} x_{r, i}^{2}} \hat{\rho}_{j, h} \hat{\rho}_{i, j} \hat{\rho}_{i, h}$
is zero.
Thus, we have
\begin{equation*}
E\left[\sum_{r} c_{r}^{\prime}\right]=\left(\frac{\gamma}{1-\gamma}\right)^4(\frac{p^{2}}{n(n-2)}+\frac{2 p(p-1)}{n(n-2)})=\left(\frac{\gamma}{1-\gamma}\right)^4\frac{3 p^{2}-2 p}{n(n-2)}
\end{equation*}
with $\alpha=\frac{\gamma}{1-\gamma}$. 
Next, the test error is calculated by summing these terms, resulting in\begin{equation*}
\begin{aligned}
&\left\{1-2\alpha+\alpha^{2} \frac{p}{n}\left(3+\frac{2}{n-2}\right)\right\}\left\|\beta_{*}\right\|^{2}+\alpha^2\frac{p}{n-2}\\&+\alpha^{3} \frac{p}{n-2}\left(\frac{p-1}{n}+1\right)+\alpha^4\frac{4p^2-2p-1+n}{n(n-2)}\end{aligned}\end{equation*}

\section{Experiment Details}
\subsection{Models}\label{model2}
Standard  CNNs. We  consider  a  simple  family  of  5-layer  CNNs,  with  four  Conv-Batch Norm-ReLU-MaxPool layers and a fully-connected output layer.  We scale the four convolutional layer widths as [$k,2k,4k,8k$].  The MaxPool is [$1, 2, 2, 8$].  For all the convolution layers, the kernel size = $3$, stride = $1$, and padding = $1$. This architecture is based on the “backbone” architecture from \cite{46}. Fork= 64, this CNN has 1558026 parameters and can reach$>90\%$ test accuracy on CIFAR-10 \citep{47} with data augmentation. The scaling of model size with $k$ is shown in "Figure 13" of \cite{6}.




\end{appendices}
\bmhead{Acknowledgments}

TLY was supported by JST, the establishment of university fellowships towards the creation of science technology innovation, Grant Number JPMJFS2125. JS was supported by Grants-in-Aid for Scientific Research (C) 22K11931.

\section*{Declarations}
\begin{itemize}
\item Funding - TLY was supported by JST, the establishment of university fellowships towards the creation of science technology innovation, Grant Number JPMJFS2125. JS was supported by Grants-in-Aid for Scientific Research (C) 22K11931.
\item Conflict of interest/Competing interests - The authors declare no conflict of interest.
\item Ethics approval - Not Applicable.
\item Consent to participate - Not Applicable.
\item Consent for publication - Not Applicable.
\item Availability of data and materials - Not Applicable.
\item Code availability - The code will be publicly available once the work is published upon the agreement of different sides.
\item Authors' contributions: 
TLY: Idea and Methodology; JS: Methodology and Supervision. All authors discussed the theoretical and experimental results and contributed to the final manuscript.
\end{itemize}


\bibliography{ref}

\end{document}

%% file: fig1.tex
\begin{tikzpicture}                 
\begin{axis}[width=8.5cm, height=5.7cm, tick align=outside, grid=both,
    grid style={line width=.1pt, draw=gray!10}, legend style={nodes={scale=0.55, transform shape}}]                        

\addplot[red!10!green,sharp plot,mark=*, mark size=0.5pt,mark options={solid}]               
coordinates                         
{                                   
 (0,1.16049851508125)
(20,1.17178963088615)
(40,1.13391982308761)
(60,1.11358255197063)
(80,1.06313482905975)
(100,1.04943632781937)
(120,0.993799920106274)
(140,0.97698780238814)
(160,0.959035345221291)
(180,0.949321875102565)
(200,0.944529945401828)
(220,0.92305326925241)
(240,0.901225750766847)
(260,0.900108712575494)
(280,0.876939225723157)
(300,0.845492934381765)
(320,0.824959117321136)
(340,0.812217749132462)
(360,0.805444638502035)
(380,0.802065489719931)
(400,0.785973490140121)
(420,0.776082918050534)
(440,0.76398118191989)
(460,0.744833843770234)
(480,0.742548140720215)
(500,0.730689208079664)
(520,0.72833033564177)
(540,0.721692560234121)
(560,0.718404975106241)
(580,0.71989463429286)
(600,0.717454690211568)
(620,0.707020113398828)
(640,0.70530946528784)
(660,0.693617553947702)
(680,0.691588903370622)
(700,0.68828886124985)
(720,0.682408720406725)
(740,0.684108436825325)
(760,0.677026430373681)
(780,0.669595449609874)
(800,0.665527913058425)
(820,0.659559952711755)
(840,0.659168177058725)
(860,0.657131132839892)
(880,0.66242056507853)
(900,0.654114896118396)
(920,0.647912030935272)
(940,0.647557817901885)
(960,0.641593692442472)
(980,0.640721081125502)
(1000,0.635503584956833)
};
\addlegendentry{$\gamma$=0.4}
\addplot[sharp plot,mark=*, mark size=0.5pt,mark options={solid},green!100!blue]
coordinates
{
(0,1.16049851508125)
(20,1.17292630763708)
(40,1.13577307596442)
(60,1.11585973471872)
(80,1.065514709572)
(100,1.05361208248325)
(120,0.993753471880195)
(140,0.976200082727804)
(160,0.953507584857387)
(180,0.942681976118748)
(200,0.93843233881467)
(220,0.911257136410019)
(240,0.884756735397574)
(260,0.882530554949101)
(280,0.855770438384959)
(300,0.815975366594209)
(320,0.790530881834445)
(340,0.77430528427565)
(360,0.764604627974677)
(380,0.760322844933323)
(400,0.739927459642425)
(420,0.724942940896944)
(440,0.710670712639668)
(460,0.68848134586655)
(480,0.685758877709907)
(500,0.671383375191855)
(520,0.666368130626452)
(540,0.65925997000476)
(560,0.653835865258649)
(580,0.653674313200387)
(600,0.648786894962158)
(620,0.636829313312095)
(640,0.633245161516235)
(660,0.621983986118083)
(680,0.62027634559083)
(700,0.614927444634161)
(720,0.60706084905156)
(740,0.608367327288849)
(760,0.600668852065717)
(780,0.592271890031683)
(800,0.586879771954692)
(820,0.581921460196413)
(840,0.582367710163822)
(860,0.580000473807037)
(880,0.585879973582867)
(900,0.577992720294133)
(920,0.571351172358635)
(940,0.570381659176701)
(960,0.564064010012786)
(980,0.562037900047964)
(1000,0.556527284923749)
};
\addlegendentry{$\gamma$=0.5}
\addplot[sharp plot,mark=*, mark size=0.5pt,mark options={solid},green!90!blue]
coordinates
{
(0,1.16049851508125)
(20,1.17373294366952)
(40,1.13726519122318)
(60,1.11807426317853)
(80,1.06831535996651)
(100,1.05896168281566)
(120,0.996847984676255)
(140,0.979538080998376)
(160,0.953434659588702)
(180,0.942378302216776)
(200,0.939606220684661)
(220,0.906893230347896)
(240,0.875994110296848)
(260,0.872929457405526)
(280,0.842653588038717)
(300,0.794134686117086)
(320,0.763808343356221)
(340,0.745148795508853)
(360,0.731854828603458)
(380,0.727272770839688)
(400,0.703048364949669)
(420,0.681315065429972)
(440,0.664854956241413)
(460,0.640240931938617)
(480,0.637886607440256)
(500,0.621651366036565)
(520,0.613494584612991)
(540,0.606565497446001)
(560,0.599274520747771)
(580,0.597068911205255)
(600,0.588642487145975)
(620,0.575903165595084)
(640,0.570275271163168)
(660,0.559882431698295)
(680,0.559118487567783)
(700,0.551592191874607)
(720,0.5412699234409)
(740,0.542177674095466)
(760,0.534532004453073)
(780,0.525673469010492)
(800,0.519565005195944)
(820,0.516848593036176)
(840,0.518655439822116)
(860,0.516166061533909)
(880,0.522366923596628)
(900,0.515894619766456)
(920,0.5093731550429)
(940,0.507374084705985)
(960,0.501171965831451)
(980,0.497701400677731)
(1000,0.492763314022807)
};
\addlegendentry{$\gamma$=0.6}
\addplot[sharp plot,mark=*, mark size=0.5pt,mark options={solid},green!80!blue]
coordinates
{
(0,1.16049851508125)
(20,1.1743347283391)
(40,1.13847956452265)
(60,1.12011525822677)
(80,1.07116324600186)
(100,1.06480135274541)
(120,1.00189267426468)
(140,0.98567493475511)
(160,0.957613492982473)
(180,0.94752634382995)
(200,0.947493149966801)
(220,0.909533185771126)
(240,0.874711041015703)
(260,0.871010846137375)
(280,0.837061572991107)
(300,0.779894748314688)
(320,0.744706205647902)
(340,0.72519890110863)
(360,0.70695650086457)
(380,0.702922418734247)
(400,0.675346274106952)
(420,0.644883576899727)
(440,0.626117602511941)
(460,0.59955242946822)
(480,0.599030973502982)
(500,0.581592302722974)
(520,0.569416435036371)
(540,0.563151945977491)
(560,0.554388304316995)
(580,0.549701578407664)
(600,0.536073298977175)
(620,0.523723212937366)
(640,0.515975124666514)
(660,0.506621422818615)
(680,0.50770031650263)
(700,0.497951925884699)
(720,0.484417663048964)
(740,0.484724677032925)
(760,0.477833088840108)
(780,0.468944136857598)
(800,0.463187655496048)
(820,0.464067468008935)
(840,0.467914341375824)
(860,0.465663318839461)
(880,0.471944063671477)
(900,0.467942346209747)
(920,0.462142225161782)
(940,0.458515245619754)
(960,0.452979900619251)
(980,0.447839161410872)
(1000,0.444588776259819)
};
\addlegendentry{$\gamma$=0.7}
\addplot[sharp plot,mark=*, mark size=0.5pt,mark options={solid},green!70!blue]
coordinates
{
(0,1.16049851508125)
(20,1.17480077302917)
(40,1.13948208217644)
(60,1.12195934616424)
(80,1.07390509538894)
(100,1.0707832482392)
(120,1.00820607207336)
(140,0.993866221733396)
(160,0.965470905954581)
(180,0.958043289117979)
(200,0.962589532542623)
(220,0.920046855625806)
(240,0.882492877112089)
(260,0.878326428422846)
(280,0.840595285365805)
(300,0.776578718786037)
(320,0.736882258767686)
(340,0.719280955308125)
(360,0.693368110222649)
(380,0.691592791480505)
(400,0.660715745072575)
(420,0.619708941395483)
(440,0.598716041637)
(460,0.570590535932012)
(480,0.5753977385035)
(500,0.557281150127807)
(520,0.538483270310791)
(540,0.532581129738809)
(560,0.522635459785065)
(580,0.514669649704251)
(600,0.493004396405246)
(620,0.483377282707765)
(640,0.473521906374539)
(660,0.465311030014377)
(680,0.469593785861149)
(700,0.457638110184022)
(720,0.43949073609422)
(740,0.438247342633678)
(760,0.43286245451944)
(780,0.424007999583825)
(800,0.420400564514433)
(820,0.4263611066756)
(840,0.433009019414267)
(860,0.431555460633438)
(880,0.437734333782123)
(900,0.437230764472574)
(920,0.432820760418351)
(940,0.426583778495114)
(960,0.422316385875251)
(980,0.415515178586034)
(1000,0.415372281735423)
};
\addlegendentry{$\gamma$=0.8}
\addplot[sharp plot,mark=*, mark size=0.5pt,mark options={solid},green!60!blue]
coordinates
{
(0,1.16049851508125)
(20,1.17517229040595)
(40,1.14032143692841)
(60,1.12361415781115)
(80,1.0764813711942)
(100,1.07672252163242)
(120,1.0153751595976)
(140,1.00368848930634)
(160,0.976848460335952)
(180,0.97453679964239)
(200,0.986577756778095)
(220,0.941146177677258)
(240,0.904144222493188)
(260,0.900161528278575)
(280,0.859658008217884)
(300,0.796776944207666)
(320,0.755281036262062)
(340,0.747152713216316)
(360,0.70828618409452)
(380,0.71573100551821)
(400,0.680135888078014)
(420,0.629120169863558)
(440,0.608790488513176)
(460,0.581497952606413)
(480,0.604217235907102)
(500,0.586066494959417)
(520,0.548217729617471)
(540,0.537070964898047)
(560,0.524817816148839)
(580,0.512186757185972)
(600,0.475900236811098)
(620,0.475352683034159)
(640,0.463165827056804)
(660,0.456674285290537)
(680,0.465544158719488)
(700,0.45152973409962)
(720,0.425290903018535)
(740,0.417950512445401)
(760,0.414510243004519)
(780,0.403731108787278)
(800,0.40516629978637)
(820,0.417630860682856)
(840,0.426892722402338)
(860,0.426808860860363)
(880,0.432769948237635)
(900,0.436252477901043)
(920,0.433832330719726)
(940,0.42280456600744)
(960,0.420015445500589)
(980,0.412275853132649)
(1000,0.416701324245732)
};
\addlegendentry{$\gamma$=0.9}
\addplot[sharp plot,mark=*, mark size=0.5pt,mark options={solid},green!50!blue]
coordinates
{
(0,1.16049851508125)
(20,1.17533114424804)
(40,1.14069147877578)
(60,1.12437600318143)
(80,1.07770101422233)
(100,1.0796429084922)
(120,1.0191949715956)
(140,1.00913357494025)
(160,0.983900192803912)
(180,0.985429797831182)
(200,1.00291009307894)
(220,0.957455325206653)
(240,0.923646978186464)
(260,0.920864077779258)
(280,0.881074184891794)
(300,0.828143362510834)
(320,0.790754269267136)
(340,0.798984847480008)
(360,0.752223317610149)
(380,0.779633094637806)
(400,0.745286696020255)
(420,0.691903524876046)
(440,0.68190598862155)
(460,0.66240876270213)
(480,0.717628796725)
(500,0.703463414955776)
(520,0.629712089958335)
(540,0.600774482449272)
(560,0.581772537016251)
(580,0.567307483224671)
(600,0.515054159227949)
(620,0.525690157284202)
(640,0.510139995200058)
(660,0.502888383163849)
(680,0.511161881904496)
(700,0.4967518791453)
(720,0.462468763897002)
(740,0.443686825617372)
(760,0.439330680525705)
(780,0.422948733786618)
(800,0.427498083029639)
(820,0.443048316867443)
(840,0.451179554704228)
(860,0.451567610596941)
(880,0.456926725759917)
(900,0.461620142222035)
(920,0.45995085206579)
(940,0.444176667943088)
(960,0.441372728336597)
(980,0.433827325544465)
(1000,0.440244092636398)
};
\addlegendentry{$\gamma$=0.95}
\addplot[sharp plot,mark=*, mark size=0.5pt,mark options={solid},green!40!blue]
coordinates
{
(0,1.16049851508125)
(20,1.17536109158707)
(40,1.14076199840518)
(60,1.12452344423656)
(80,1.07793939006809)
(100,1.0802219332149)
(120,1.01997448650781)
(140,1.01026281327748)
(160,0.985425158064205)
(180,0.987851260564544)
(200,1.00660302072528)
(220,0.961336689433649)
(240,0.928579938749387)
(260,0.926261513073067)
(280,0.887068832195919)
(300,0.837735270585985)
(320,0.802341863068226)
(340,0.816857919591299)
(360,0.769189576782055)
(380,0.805611078204547)
(400,0.775164559994728)
(420,0.721999587960828)
(440,0.718385599720215)
(460,0.703138526136659)
(480,0.77312689028448)
(500,0.76181347978721)
(520,0.672806960994679)
(540,0.634917269312894)
(560,0.610951521792369)
(580,0.596278061544686)
(600,0.538377544021586)
(620,0.552137560772846)
(640,0.534682646827756)
(660,0.52552408849229)
(680,0.531947791177537)
(700,0.517768305745786)
(720,0.480981956673993)
(740,0.457493037080853)
(760,0.452003623973095)
(780,0.433170717012384)
(800,0.437984472901329)
(820,0.453907400820441)
(840,0.461088757119356)
(860,0.461463461679877)
(880,0.466451002353988)
(900,0.471107643568173)
(920,0.469444355640162)
(940,0.452254076266596)
(960,0.449224418512865)
(980,0.441757165847592)
(1000,0.448329217539823)
};
\addlegendentry{$\gamma$=0.96}
\addplot[sharp plot,mark=*, mark size=0.5pt,mark options={solid},green!30!blue]
coordinates
{
(0,1.16049851508125)
(20,1.17539047138266)
(40,1.1408314168218)
(60,1.12466928970252)
(80,1.07817591969149)
(100,1.08079915203015)
(120,1.0207587892267)
(140,1.01140520752194)
(160,0.986989480907816)
(180,0.990359486897927)
(200,1.01045358436722)
(220,0.965458712869697)
(240,0.933935364674187)
(260,0.932198693110722)
(280,0.893853980704917)
(300,0.848913000580324)
(320,0.816220782346301)
(340,0.839069990088344)
(360,0.791337130434015)
(380,0.840838428149449)
(400,0.819018220910184)
(420,0.767130186071567)
(440,0.774591269717989)
(460,0.766465021459233)
(480,0.859711177558735)
(500,0.853102295331733)
(520,0.742886608458674)
(540,0.691143795412719)
(560,0.656406118319569)
(580,0.640991163578479)
(600,0.575311478384159)
(620,0.592277952645587)
(640,0.571412780947533)
(660,0.557919397790764)
(680,0.560715807643206)
(700,0.547090030875116)
(720,0.507164918180887)
(740,0.477088599891464)
(760,0.469592650770416)
(780,0.447355354557486)
(800,0.452007905510307)
(820,0.468079471085238)
(840,0.473830285928266)
(860,0.474120874001722)
(880,0.478505590178554)
(900,0.482921384112666)
(920,0.481154482553222)
(940,0.462279243530332)
(960,0.458884184350039)
(980,0.451478043653525)
(1000,0.458026738241975)
};
\addlegendentry{$\gamma$=0.97}
\addplot[sharp plot,mark=*, mark size=0.5pt,mark options={solid},green!20!blue]
coordinates
{
(0,1.16049851508125)
(20,1.17541929959908)
(40,1.14089975894086)
(60,1.1248135596996)
(80,1.07841060738243)
(100,1.08137451941898)
(120,1.02154772345792)
(140,1.01256066989668)
(160,0.988593727022301)
(180,0.992957273377084)
(200,1.01446946407591)
(220,0.969839367467963)
(240,0.939756311508967)
(260,0.938746057993776)
(280,0.90156992664639)
(300,0.861987699990403)
(320,0.832954013468792)
(340,0.86717552029058)
(360,0.82098289126443)
(380,0.890496525059758)
(400,0.887977398104346)
(420,0.841139969174318)
(440,0.870595382786179)
(460,0.877458551715887)
(480,1.01390447590058)
(500,1.01472757339897)
(520,0.874406341519166)
(540,0.799159004222626)
(560,0.735258960964434)
(580,0.71608164377544)
(600,0.638312254026417)
(620,0.657807497013755)
(640,0.629817453965482)
(660,0.606175454346144)
(680,0.601829152954311)
(700,0.589396798487393)
(720,0.545252018927419)
(740,0.505381785478286)
(760,0.494336161230258)
(780,0.4672173081491)
(800,0.470904600913)
(820,0.486738768369246)
(840,0.490365703404094)
(860,0.490472475610257)
(880,0.493854172245535)
(900,0.49771349373133)
(920,0.495676251347533)
(940,0.474754516503813)
(960,0.47080213028268)
(980,0.463403488171184)
(1000,0.469669867167879)
};
\addlegendentry{$\gamma$=0.98}
\addplot[sharp plot,mark=*, mark size=0.5pt,mark options={solid},green!10!blue]
coordinates
{
(0,1.16049851508125)
(20,1.17544759160811)
(40,1.14096704895955)
(60,1.12495627427672)
(80,1.07864345840854)
(100,1.08194799286307)
(120,1.02234113914606)
(140,1.01372912196852)
(160,0.990238499698575)
(180,0.995647586451779)
(200,1.01865895099955)
(220,0.974498733883094)
(240,0.946092105731194)
(260,0.945987238747142)
(280,0.91039326960124)
(300,0.877356133341889)
(320,0.853308682214478)
(340,0.903604270525342)
(360,0.862116835066876)
(380,0.964610202099652)
(400,1.00893340448817)
(420,0.984579678497494)
(440,1.07159072884882)
(460,1.13143739030526)
(480,1.38061193577647)
(500,1.39144252712947)
(520,1.20560255498342)
(540,1.08087747952275)
(560,0.904067982537843)
(580,0.863950510819197)
(600,0.762169728758941)
(620,0.780416139163499)
(640,0.73324591928949)
(660,0.683257452833791)
(680,0.663879782177236)
(700,0.653982786640372)
(720,0.603434764059831)
(740,0.547628543683869)
(760,0.530144318143331)
(780,0.495708467585516)
(800,0.496840350107742)
(820,0.511731349482919)
(840,0.512195902071293)
(860,0.511969828276871)
(880,0.513629826390885)
(900,0.516420554980371)
(920,0.513852354260311)
(940,0.490383367699977)
(960,0.485605310234414)
(980,0.478094864299912)
(1000,0.483701311480763)
};
\addlegendentry{$\gamma$=0.99}
\addplot[sharp plot,mark=*, mark size=0.5pt,mark options={solid},green!1!blue]
coordinates
{
(0,1.16049851508125)
(20,1.17546154118336)
(40,1.14100030680284)
(60,1.12502705452365)
(80,1.07875919706615)
(100,1.08223400684994)
(120,1.02273948228147)
(140,1.01431819711236)
(160,0.991076282539241)
(180,0.997028416483866)
(200,1.02082156967789)
(220,0.976939863511695)
(240,0.949470480988334)
(260,0.949898508704414)
(280,0.915288286051897)
(300,0.88605554767276)
(320,0.86516461731493)
(340,0.926105605947158)
(360,0.889128418898869)
(380,1.0169700161318)
(400,1.11012280138449)
(420,1.12429927257008)
(440,1.29161552010692)
(460,1.46688779243873)
(480,1.88784231774804)
(500,1.91811437123901)
(520,1.67499290903682)
(540,1.47811635007004)
(560,1.0917776010174)
(580,1.00892023072966)
(600,0.88043489970304)
(620,0.892596267411548)
(640,0.820724128805189)
(660,0.741561462336966)
(680,0.708207998150155)
(700,0.700484449810371)
(720,0.64504425919273)
(740,0.576925706631226)
(760,0.554326117052144)
(780,0.514775138972218)
(800,0.513536188306967)
(820,0.527497839233269)
(840,0.525818846171157)
(860,0.525342679779045)
(880,0.525694020942914)
(900,0.527653864593699)
(920,0.524675720757315)
(940,0.499681649687632)
(960,0.494356274810377)
(980,0.486711061645263)
(1000,0.491793043771835)
};
\addlegendentry{$\gamma$=0.995}
\end{axis}                         
\end{tikzpicture}

%% file: fig5.tex
\begin{tikzpicture}                 
\begin{axis}[width=8.5cm, height=4.0cm, tick align=outside, grid=both,
    grid style={line width=.1pt, draw=gray!10},
    legend style={nodes={scale=1.0, transform shape}}]                        

\addplot[red,sharp plot,mark=*, mark size=0.5pt,mark options={solid}]               
coordinates
{

(21,34.5685245390091)
(41,20.1794249089152)
(61,14.922704654614)
(81,12.1418794561723)
(101,10.4004366444047)
(121,9.19780954586738)
(141,8.31231717655369)
(161,7.630126946363)
(181,7.08654341579919)
(201,6.64196367843728)
(221,6.27072762689496)
(241,5.95544480625888)
(261,5.68389147405186)
(281,5.44721207316602)
(301,5.23882587923864)
(321,5.05373528025048)
(341,4.88807305830646)
(361,4.73879731703578)
(381,4.60348061341682)
(401,4.48016091543522)
(421,4.36723415619087)
(441,4.26337539807236)
(461,4.16748006666512)
(481,4.0786195152165)
(501,3.9960069870247)
(521,3.91897123304028)
(541,3.84693584087923)
(561,3.77940287727135)
(581,3.71593982488287)
(601,3.65616906138645)
(621,3.5997593192599)
(641,3.54641870261693)
(661,3.4958889381912)
(681,3.44794061214627)
(701,3.40236920007026)
(721,3.3589917395038)
(741,3.31764402629238)
(761,3.27817824055622)
(781,3.24046092701336)
(801,3.20437126914324)
(821,3.16979960824763)
(841,3.13664616759795)
(861,3.10481994911258)
(881,3.07423777580432)
(901,3.04482345789565)
(921,3.01650706426263)
(941,2.98922428392284)
(961,2.96291586477613)
(981,2.93752711884959)
(1001,2.91300748498093)

};
\addlegendentry{$(1+\sqrt{\frac{p}{n}})^2$}
\addplot[green,sharp plot,mark=*, mark size=0.5pt,mark options={solid}]               
coordinates
{

(21,32.2749267890736)
(41,19.9332078420058)
(61,14.3659543437588)
(81,11.6502146160112)
(101,10.0268083248173)
(121,8.58182433157757)
(141,8.02954757071722)
(161,7.43417547423058)
(181,6.81726085688279)
(201,6.38024228406071)
(221,6.05416014203431)
(241,5.80644231465901)
(261,5.51266437827659)
(281,5.26614649548988)
(301,5.05854816483826)
(321,4.80901044047887)
(341,4.63485621645857)
(361,4.62192204773239)
(381,4.45302206598076)
(401,4.30438287026595)
(421,4.21553761326961)
(441,4.13335963398548)
(461,4.0275035412547)
(481,3.9896129603862)
(501,3.92683562193881)
(521,3.8367721364167)
(541,3.7386477408464)
(561,3.70691134707306)
(581,3.64889358809258)
(601,3.63304962724783)
(621,3.56003197355079)
(641,3.47874494920839)
(661,3.44163973613154)
(681,3.39824007713766)
(701,3.34211914273617)
(721,3.30629168291437)
(741,3.2654483917825)
(761,3.21912505686797)
(781,3.21843209821225)
(801,3.1773744334532)
(821,3.13435947077136)
(841,3.11392976619036)
(861,3.06666468622255)
(881,3.0384637192976)
(901,3.02076499722494)
(921,3.02413281908455)
(941,2.981160078074)
(961,2.94559352919604)
(981,2.9292785682422)
(1001,2.91030314609965)

};
\addlegendentry{largest eigenvalue of M}
\end{axis}
\end{tikzpicture}

%% file: fig2.tex
\begin{tikzpicture}                 
\begin{axis}[width=8.5cm, height=5.7cm, tick align=outside, grid=both,
    grid style={line width=.1pt, draw=gray!10},
    legend style={nodes={scale=0.55, transform shape}}]                        

\addplot[green,sharp plot,mark=*, mark size=0.5pt,mark options={solid}]               
coordinates
{
(0,0.9195)
(100,0.3879)
(200,0.1887)
(300,0.1536)
(400,0.1281)
(500,0.1015)
(600,0.0945)
(700,0.0844)
(800,0.0779)
(900,0.0744)
(1000,0.0677)
(1100,0.0668)
(1200,0.0638)
(1300,0.061)
(1400,0.0592)
(1500,0.0568)
(1600,0.0578)
(1700,0.0563)
(1800,0.0543)
(1900,0.0551)
(2000,0.0556)
};
\addlegendentry{$p$=100}
\addplot[green!90!blue,sharp plot,mark=*, mark size=0.5pt,mark options={solid}]               
coordinates
{(0,1.0258)
(100,0.6345)
(200,0.3516)
(300,0.2588)
(400,0.1786)
(500,0.1433)
(600,0.1193)
(700,0.1051)
(800,0.1006)
(900,0.0925)
(1000,0.0834)
(1100,0.0839)
(1200,0.0823)
(1300,0.0752)
(1400,0.0699)
(1500,0.0661)
(1600,0.0619)
(1700,0.0597)
(1800,0.0587)
(1900,0.0565)
(2000,0.0575)
};
\addlegendentry{$p$=200}
\addplot[green!80!blue,sharp plot,mark=*, mark size=0.5pt,mark options={solid}]               
coordinates
{
(0,0.9897)
(100,0.7684)
(200,0.4712)
(300,0.3571)
(400,0.2904)
(500,0.2416)
(600,0.2029)
(700,0.182)
(800,0.1735)
(900,0.1531)
(1000,0.1438)
(1100,0.1228)
(1200,0.1109)
(1300,0.1048)
(1400,0.1008)
(1500,0.0963)
(1600,0.0944)
(1700,0.0939)
(1800,0.0897)
(1900,0.0857)
(2000,0.0842)
};
\addlegendentry{$p$=300}
\addplot[green!70!blue,sharp plot,mark=*, mark size=0.5pt,mark options={solid}]               
coordinates
{
(0,0.978)
(100,0.8107)
(200,0.6193)
(300,0.5424)
(400,0.3943)
(500,0.3107)
(600,0.2631)
(700,0.215)
(800,0.1931)
(900,0.1685)
(1000,0.1696)
(1100,0.156)
(1200,0.1403)
(1300,0.131)
(1400,0.124)
(1500,0.118)
(1600,0.1128)
(1700,0.1081)
(1800,0.0982)
(1900,0.0969)
(2000,0.0924)
};
\addlegendentry{$p$=400}
\addplot[green!60!blue,sharp plot,mark=*, mark size=0.5pt,mark options={solid}]               
coordinates
{
(0,0.9368)
(100,0.8582)
(200,0.7589)
(300,0.5959)
(400,0.4979)
(500,0.3947)
(600,0.3621)
(700,0.3057)
(800,0.2481)
(900,0.2072)
(1000,0.1934)
(1100,0.1749)
(1200,0.1652)
(1300,0.1429)
(1400,0.1379)
(1500,0.1237)
(1600,0.1185)
(1700,0.1153)
(1800,0.1141)
(1900,0.1055)
(2000,0.0977)
};
\addlegendentry{$p$=500}
\addplot[green!50!blue,sharp plot,mark=*, mark size=0.5pt,mark options={solid}]               
coordinates
{
(0,1.0526)
(100,0.9468)
(200,0.806)
(300,0.6551)
(400,0.5097)
(500,0.4539)
(600,0.4297)
(700,0.3892)
(800,0.3357)
(900,0.2706)
(1000,0.2331)
(1100,0.2092)
(1200,0.1812)
(1300,0.1716)
(1400,0.1679)
(1500,0.1558)
(1600,0.1492)
(1700,0.1451)
(1800,0.1392)
(1900,0.131)
(2000,0.127)
};
\addlegendentry{$p$=600}
\addplot[green!40!blue,sharp plot,mark=*, mark size=0.5pt,mark options={solid}]               
coordinates
{
(0,1.0291)
(100,0.9319)
(200,0.8575)
(300,0.7693)
(400,0.6726)
(500,0.4973)
(600,0.4627)
(700,0.4083)
(800,0.3504)
(900,0.2728)
(1000,0.268)
(1100,0.2483)
(1200,0.2309)
(1300,0.2226)
(1400,0.2146)
(1500,0.2004)
(1600,0.1883)
(1700,0.1759)
(1800,0.1678)
(1900,0.165)
(2000,0.1647)
};
\addlegendentry{$p$=700}
\addplot[green!30!blue,sharp plot,mark=*, mark size=0.5pt,mark options={solid}]               
coordinates
{
(0,1.0008)
(100,0.9277)
(200,0.8452)
(300,0.7509)
(400,0.6401)
(500,0.5704)
(600,0.4795)
(700,0.4153)
(800,0.3593)
(900,0.3167)
(1000,0.2892)
(1100,0.2709)
(1200,0.2524)
(1300,0.2336)
(1400,0.2226)
(1500,0.2083)
(1600,0.1919)
(1700,0.1824)
(1800,0.1724)
(1900,0.1678)
(2000,0.1607)
};
\addlegendentry{$p$=800}
\addplot[green!20!blue,sharp plot,mark=*, mark size=0.5pt,mark options={solid}]               
coordinates
{
(0,0.92)
(100,0.8326)
(200,0.7276)
(300,0.6764)
(400,0.6096)
(500,0.5693)
(600,0.5007)
(700,0.4771)
(800,0.4404)
(900,0.3943)
(1000,0.3601)
(1100,0.3344)
(1200,0.305)
(1300,0.2773)
(1400,0.261)
(1500,0.2401)
(1600,0.2163)
(1700,0.2132)
(1800,0.197)
(1900,0.1942)
(2000,0.1843)
};
\addlegendentry{$p$=900}
\addplot[green!10!blue,sharp plot,mark=*, mark size=0.5pt,mark options={solid}]               
coordinates
{
(0,1.0012)
(100,0.8419)
(200,0.7814)
(300,0.7381)
(400,0.679)
(500,0.6129)
(600,0.555)
(700,0.4748)
(800,0.4236)
(900,0.3837)
(1000,0.3486)
(1100,0.3104)
(1200,0.3014)
(1300,0.2787)
(1400,0.2736)
(1500,0.255)
(1600,0.2273)
(1700,0.2181)
(1800,0.2045)
(1900,0.1946)
(2000,0.1824)
};
\addlegendentry{$p$=1000}
\end{axis}                         
\end{tikzpicture}

%% file: fig3.tex
\begin{tikzpicture}                 
\begin{axis}[width=8.5cm, height=5.7cm, tick align=outside, grid=both,ytick={0,0.10,0.20},
yticklabel style={
/pgf/number format/.cd,
precision=2,
/tikz/.cd
},grid style={line width=.1pt, draw=gray!10},
legend style={nodes={scale=0.55, transform shape}}]                        

\addplot[red!30!green,sharp plot,mark=*, mark size=0.5pt,mark options={solid}]               
coordinates
{

(0,0.1)
(20,0.0853060171297688)
(40,0.0825127442153752)
(60,0.0831981048749465)
(80,0.0832333728274296)
(100,0.0827234758012853)
(120,0.0826298885788982)
(140,0.0824570429325008)
(160,0.082300881946359)
(180,0.0824224048628002)
(200,0.0824479569333002)
(220,0.0825130654222381)
(240,0.0826314086808118)
(260,0.0827339082911386)
(280,0.0826281629227839)
(300,0.0825400430842446)
(320,0.0827395034530789)
(340,0.082724759803401)
(360,0.0827138905452272)
(380,0.0828025028457103)
(400,0.0828233733682036)
(420,0.0827601798376975)
(440,0.0826874441750152)
(460,0.0826131791138036)
(480,0.082685929823086)
(500,0.0826787037718224)
(520,0.0825966582661664)
(540,0.08265761877349)
(560,0.0826683976844458)
(580,0.0827318238868668)
(600,0.082708894637253)
(620,0.0827083957918392)
(640,0.0827459576284482)
(660,0.0827498598572773)
(680,0.0827427682071361)
(700,0.0827436054404406)
(720,0.0827293876666827)
(740,0.0827333324246295)
(760,0.082701318606978)
(780,0.0827343409266905)
(800,0.0826919725457115)
(820,0.0826747442140806)
(840,0.0826580117095597)
(860,0.0826395148827996)
(880,0.0826553968623572)
(900,0.0826284642568166)
(920,0.0826250585062453)
(940,0.0826283711578817)
(960,0.0826430451046237)
(980,0.0826338093330638)
(1000,0.0826515203935585)

};
\addlegendentry{$\gamma=0.008$}

\addplot[red!20!green,sharp plot,mark=*, mark size=0.5pt,mark options={solid}]               
coordinates
{

(0,0.1)
(20,0.14740420393251)
(40,0.0780698590689947)
(60,0.076015579508333)
(80,0.0539793139989711)
(100,0.0515164437756727)
(120,0.0494918589819895)
(140,0.0485749871135026)
(160,0.0469309599808851)
(180,0.0461537496977059)
(200,0.0457100631836578)
(220,0.0452061924382715)
(240,0.0452583261671991)
(260,0.0450766428208715)
(280,0.0443666005493421)
(300,0.0439733118048734)
(320,0.044138774001001)
(340,0.0438591860319199)
(360,0.0433276467376139)
(380,0.0436026537626445)
(400,0.0436546376759517)
(420,0.0436982718503236)
(440,0.0434029925103252)
(460,0.0430238707033013)
(480,0.0431874972846383)
(500,0.0428849004966992)
(520,0.042782075486217)
(540,0.0426853083937331)
(560,0.0426014503086398)
(580,0.0425560193760251)
(600,0.042454232537877)
(620,0.0424854520391815)
(640,0.0426224256935415)
(660,0.0425928960501808)
(680,0.0425032738965426)
(700,0.0426276161186455)
(720,0.0425435785565544)
(740,0.0425537574249374)
(760,0.0425221795468219)
(780,0.042412553048408)
(800,0.0423374571623599)
(820,0.0422430796518781)
(840,0.0421893379966542)
(860,0.0420036891255626)
(880,0.0419588821606965)
(900,0.0418718617779843)
(920,0.0418505490597334)
(940,0.0419128930728488)
(960,0.0419201810999637)
(980,0.0419011817554095)
(1000,0.0418965850186704)

};
\addlegendentry{$\gamma=0.3$}
\addplot[red!10!green,sharp plot,mark=*, mark size=0.5pt,mark options={solid}]               
coordinates
{

(0,0.1)
(20,0.15846197444657)
(40,0.0896849796172417)
(60,0.0902775584995544)
(80,0.0549383717289696)
(100,0.0526390617666998)
(120,0.0503529357844224)
(140,0.0488544799415676)
(160,0.0469346287286776)
(180,0.0458450510043074)
(200,0.0451858855197233)
(220,0.0446136533499534)
(240,0.0445758266457281)
(260,0.0443238855097892)
(280,0.0434626263292801)
(300,0.0430884860004223)
(320,0.043073252698219)
(340,0.0427672525101102)
(360,0.0419489622790042)
(380,0.0422276986708345)
(400,0.0422418681680964)
(420,0.0422842005520949)
(440,0.0419058618608324)
(460,0.0415127254738247)
(480,0.0416154800795553)
(500,0.0412516139771933)
(520,0.0411766847946802)
(540,0.0410206251537878)
(560,0.0408571015906173)
(580,0.0406984580448338)
(600,0.0405763924497998)
(620,0.0406035214810241)
(640,0.040749715252647)
(660,0.040737819039339)
(680,0.0406244393894333)
(700,0.0407610952909788)
(720,0.04065534930294)
(740,0.040665519276963)
(760,0.040641440630135)
(780,0.0404583038549385)
(800,0.0403589504779993)
(820,0.0402701894295375)
(840,0.0402190804250719)
(860,0.0399911148037267)
(880,0.0399344199215313)
(900,0.0398478167769948)
(920,0.0398046130941963)
(940,0.0398723033588253)
(960,0.0398651569457014)
(980,0.0398646783160095)
(1000,0.0398378349249718)

};
\addlegendentry{$\gamma=0.4$}
\addplot[green!100!blue,sharp plot,mark=*, mark size=0.5pt,mark options={solid}]               
coordinates
{

(0,0.1)
(20,0.16698721315596)
(40,0.10200279557429)
(60,0.107210180926351)
(80,0.0566124565767125)
(100,0.0545809387855582)
(120,0.0520914246091119)
(140,0.0498672600200328)
(160,0.0477027258135912)
(180,0.0462977692717761)
(200,0.0454039382111155)
(220,0.0448344031399939)
(240,0.0447236547873072)
(260,0.044391723480437)
(280,0.0434127368324654)
(300,0.0430918868694462)
(320,0.0429076314009095)
(340,0.0425622307244252)
(360,0.0414451900618526)
(380,0.0417410225844998)
(400,0.0417189460602728)
(420,0.0417522568957457)
(440,0.0412724731772124)
(460,0.0408552963421468)
(480,0.0408636413278482)
(500,0.0404462899126736)
(520,0.0403942978107959)
(540,0.0401661905187285)
(560,0.0399455115880843)
(580,0.0396331409596284)
(600,0.0394896835704645)
(620,0.0395030405674979)
(640,0.0396654033972895)
(660,0.0396597008307562)
(680,0.039504472541611)
(700,0.0396377792984117)
(720,0.0395129583321898)
(740,0.0395153516201048)
(760,0.0394977039994687)
(780,0.0392387335425866)
(800,0.0391130326427021)
(820,0.0390272022646715)
(840,0.0389795982952995)
(860,0.0387103378138907)
(880,0.0386386771396114)
(900,0.0385504219660741)
(920,0.0384780040210775)
(940,0.0385460881623991)
(960,0.0385204474324149)
(980,0.0385366150434069)
(1000,0.0384846918892389)

};
\addlegendentry{$\gamma=0.5$}
\addplot[green!90!blue,sharp plot,mark=*, mark size=0.5pt,mark options={solid}]               
coordinates
{

(0,0.1)
(20,0.179376177938662)
(40,0.127281746026361)
(60,0.147463683759845)
(80,0.0620698558335792)
(100,0.061000946681666)
(120,0.0583542170822608)
(140,0.0539568977167842)
(160,0.051402071587826)
(180,0.049262671841431)
(200,0.0477687666210768)
(220,0.0475192898571931)
(240,0.047260270770804)
(260,0.046773318619801)
(280,0.0456332804212354)
(300,0.045504047906565)
(320,0.0449507204651824)
(340,0.0445345433424156)
(360,0.0427942556744326)
(380,0.043176088017863)
(400,0.0430771210648582)
(420,0.0430555741284651)
(440,0.0423396710792076)
(460,0.041822508384572)
(480,0.041464704265152)
(500,0.0409633483987481)
(520,0.0409293315847255)
(540,0.0404878231917982)
(560,0.0402620085478094)
(580,0.0394460209238164)
(600,0.0392390308314184)
(620,0.0391803162020083)
(640,0.0393974924954891)
(660,0.0393479621464059)
(680,0.0390202048227662)
(700,0.0390800874513247)
(720,0.0389272590408901)
(740,0.0388607383671756)
(760,0.0388403647687971)
(780,0.0383992208894698)
(800,0.0381990744457639)
(820,0.0380997691596561)
(840,0.0380560418186842)
(860,0.0376886500155733)
(880,0.0375699500379684)
(900,0.0374609154775817)
(920,0.0372805633688309)
(940,0.0373214644477146)
(960,0.0372359733353157)
(980,0.0372739334483059)
(1000,0.0371625551292824)

};
\addlegendentry{$\gamma=0.7$}
\addplot[green!80!blue,sharp plot,mark=*, mark size=0.5pt,mark options={solid}]               
coordinates
{

(0,0.1)
(20,0.184038750913313)
(40,0.139920888949772)
(60,0.170483871653845)
(80,0.0663653070830012)
(100,0.06624488288926)
(120,0.0639317773088399)
(140,0.0578027886285182)
(160,0.0551900487809241)
(180,0.052599428367372)
(200,0.0506735763992971)
(220,0.0509216874645543)
(240,0.0504994242521086)
(260,0.0500043332909755)
(280,0.0488181020346552)
(300,0.0488171122937911)
(320,0.0479886235027907)
(340,0.0475933724603148)
(360,0.0455527827119086)
(380,0.0460162450385874)
(400,0.0458213345209866)
(420,0.0457227534094249)
(440,0.044885762574733)
(460,0.0442559615918997)
(480,0.0435021586008579)
(500,0.0429554981327134)
(520,0.0428926679322036)
(540,0.0422825724846535)
(560,0.0421445608885581)
(580,0.0408729852310884)
(600,0.0406059063334017)
(620,0.0404565162769101)
(640,0.0407115721627534)
(660,0.0405783608651706)
(680,0.0400658675364527)
(700,0.0400177353665271)
(720,0.0398627293151897)
(740,0.0396850935841221)
(760,0.0396349694367879)
(780,0.0390575195974184)
(800,0.0387839599721946)
(820,0.0386556802981484)
(840,0.0385998581479548)
(860,0.0381492764563992)
(880,0.0379855985325155)
(900,0.0378448925886324)
(920,0.0375605516851965)
(940,0.0375534855475578)
(960,0.037413154765451)
(980,0.0374493955592324)
(1000,0.0372966862765753)

};
\addlegendentry{$\gamma=0.8$}
\addplot[green!70!blue,sharp plot,mark=*, mark size=0.5pt,mark options={solid}]               
coordinates
{

(0,0.1)
(20,0.187996936562796)
(40,0.152464989434895)
(60,0.195492014279175)
(80,0.0726057232577284)
(100,0.0742985477893006)
(120,0.0735625666661752)
(140,0.0649923535891486)
(160,0.0628183593617139)
(180,0.0598355385819384)
(200,0.0574047224177522)
(220,0.0589396573263065)
(240,0.0580828127658083)
(260,0.0579230310911532)
(280,0.0568691453820517)
(300,0.0569391026057867)
(320,0.0556750948760573)
(340,0.0554408392897706)
(360,0.0533151984164424)
(380,0.0539245469743003)
(400,0.0532569780286512)
(420,0.0529873560466372)
(440,0.0520266082970927)
(460,0.051112833991995)
(480,0.0494505592547121)
(500,0.0487586402968508)
(520,0.0485423058213996)
(540,0.047703214176045)
(560,0.0477205348969303)
(580,0.045539934608694)
(600,0.0451232628279973)
(620,0.0447384673129052)
(640,0.0449972486461034)
(660,0.0446131432821538)
(680,0.0436993418211648)
(700,0.0433506466465091)
(720,0.0432250273018662)
(740,0.0427419887731585)
(760,0.0425637048246954)
(780,0.0417442445700149)
(800,0.0413024554887259)
(820,0.0410961011577959)
(840,0.0409638440730124)
(860,0.0403260846682139)
(880,0.0400504067962463)
(900,0.0398237844579299)
(920,0.0393704107457136)
(940,0.0392415705023606)
(960,0.0389969242005285)
(980,0.0390058745314577)
(1000,0.0387687788014448)

};
\addlegendentry{$\gamma=0.9$}
\addplot[green!60!blue,sharp plot,mark=*, mark size=0.5pt,mark options={solid}]               
coordinates
{

(0,0.1)
(20,0.189760664709485)
(40,0.158692070572466)
(60,0.208818326800549)
(80,0.0769425004458213)
(100,0.0802797204281533)
(120,0.0818458575051349)
(140,0.0719436071661372)
(160,0.0708858378979951)
(180,0.0681444698390222)
(200,0.0657885453854841)
(220,0.069236079478748)
(240,0.0679813156728961)
(260,0.0688887582419841)
(280,0.0686343363077813)
(300,0.0685289037651578)
(320,0.0671785372661157)
(340,0.0671360701551608)
(360,0.0657088467617498)
(380,0.0665696897329069)
(400,0.0645844538850473)
(420,0.0642747270344367)
(440,0.0630826709850536)
(460,0.0617625983395957)
(480,0.0588791890475047)
(500,0.0577851806399692)
(520,0.0572309119431282)
(540,0.0563009117630199)
(560,0.056173144498707)
(580,0.0528088761632076)
(600,0.0521008650658062)
(620,0.0513621005406682)
(640,0.0514605909677635)
(660,0.0506356951020051)
(680,0.0492397251946962)
(700,0.0483780809370008)
(720,0.0482920002839558)
(740,0.0474357380387675)
(760,0.0469854556483977)
(780,0.0458990690507659)
(800,0.0452269345260786)
(820,0.0449199089616667)
(840,0.0446226723619325)
(860,0.0437197522925116)
(880,0.0432724953546267)
(900,0.0429166941047376)
(920,0.0423535491580584)
(940,0.0420760262471475)
(960,0.0417121315149555)
(980,0.0416824807402527)
(1000,0.041315313033584)

};
\addlegendentry{$\gamma=0.95$}
\addplot[green!50!blue,sharp plot,mark=*, mark size=0.5pt,mark options={solid}]               
coordinates
{

(0,0.1)
(20,0.190430909287158)
(40,0.161173895954182)
(60,0.214318764113846)
(80,0.0790163929281322)
(100,0.0832737374203236)
(120,0.0864629400055774)
(140,0.0762093250575677)
(160,0.0762334780329989)
(180,0.0740529371161625)
(200,0.0721902848103914)
(220,0.077490771299138)
(240,0.0762289121438992)
(260,0.0786213526930401)
(280,0.0797732951069253)
(300,0.0794858479593396)
(320,0.0785929684919292)
(340,0.0786661569420435)
(360,0.0785803945278947)
(380,0.0798706427117113)
(400,0.076024964098633)
(420,0.0759759072980606)
(440,0.0743104860615832)
(460,0.0726765020328626)
(480,0.0686829586389821)
(500,0.0669871010230631)
(520,0.0660357298555749)
(540,0.065108624299568)
(560,0.0644287808401741)
(580,0.0598329783967421)
(600,0.0587481364872032)
(620,0.0576458382793361)
(640,0.0574837456960635)
(660,0.0561757065529808)
(680,0.0544208259877801)
(700,0.0530096895298207)
(720,0.0529217946080047)
(740,0.0518271682652235)
(760,0.0510622544017853)
(780,0.0497383777979777)
(800,0.0488383060299936)
(820,0.0484679515786298)
(840,0.0479903865356205)
(860,0.0468383379147406)
(880,0.0462306066634547)
(900,0.0457596563359028)
(920,0.0451525230512454)
(940,0.0447458360324672)
(960,0.0442748650235244)
(980,0.0442220939889317)
(1000,0.0437308415442416)

};
\addlegendentry{$\gamma=0.97$}
\addplot[green!40!blue,sharp plot,mark=*, mark size=0.5pt,mark options={solid}]               
coordinates
{

(0,0.1)
(20,0.190758909756351)
(40,0.162412845365881)
(60,0.217107646626908)
(80,0.0801437120746015)
(100,0.0849418249143178)
(120,0.0891973103379901)
(140,0.0788891756543085)
(160,0.0797736757364152)
(180,0.0781680720472242)
(200,0.0768930644986345)
(220,0.0838803133291315)
(240,0.0828815050617797)
(260,0.0869046493564718)
(280,0.0898561423361312)
(300,0.0895655342279722)
(320,0.0894841940228693)
(340,0.0897026030042465)
(360,0.0914143586491674)
(380,0.0933371521246377)
(400,0.0873135677044621)
(420,0.0877617057449893)
(440,0.0854134312142659)
(460,0.0836273506632806)
(480,0.0786213318358674)
(500,0.0762000797002167)
(520,0.0748522117835563)
(540,0.073929852859629)
(560,0.0724099795461149)
(580,0.066418000269638)
(600,0.0649093442865166)
(620,0.0634111779217567)
(640,0.0629668228018285)
(660,0.061167839719132)
(680,0.0591672141896547)
(700,0.057219847223324)
(720,0.0570854092667656)
(740,0.0558441539726435)
(760,0.0547702152317602)
(780,0.053205534580097)
(800,0.0520720806888838)
(820,0.0516777998080732)
(840,0.0510216296814351)
(860,0.0496414693381929)
(880,0.0488872859794874)
(900,0.0483191501962268)
(920,0.0476817830021295)
(940,0.0471487883299494)
(960,0.046575697601968)
(980,0.0465123653616582)
(1000,0.0459148987642635)

};
\addlegendentry{$\gamma=0.98$}
\addplot[green!30!blue,sharp plot,mark=*, mark size=0.5pt,mark options={solid}]               
coordinates
{

(0,0.1)
(20,0.191082296596458)
(40,0.163650478789501)
(60,0.219923181284135)
(80,0.0813389242507627)
(100,0.0867438271323311)
(120,0.092297130543742)
(140,0.0820805084629513)
(160,0.0841955382244411)
(180,0.0835714220894666)
(200,0.0834311792940703)
(220,0.0934319381163688)
(240,0.0933921876050292)
(260,0.10085360515476)
(280,0.108395243602575)
(300,0.108885905987844)
(320,0.11128799888753)
(340,0.112272155861814)
(360,0.118983280828087)
(380,0.122984311032255)
(400,0.11161283678104)
(420,0.113699400607253)
(440,0.109477885509483)
(460,0.108083198011842)
(480,0.101029374236471)
(500,0.0968407178771249)
(520,0.094711737397382)
(540,0.0937911040486979)
(560,0.0897622811114334)
(580,0.0798024451964587)
(600,0.0773029441240146)
(620,0.0747181649476502)
(640,0.0737194840906903)
(660,0.0708567085864198)
(680,0.0685769399929225)
(700,0.0655653392616183)
(720,0.0651996567439418)
(740,0.0637432465628991)
(760,0.0620802756396008)
(780,0.0599115006998356)
(800,0.0582391956484262)
(820,0.0578695513161456)
(840,0.0568371060503174)
(860,0.0550007122657604)
(880,0.0539485679363314)
(900,0.0532017232564261)
(920,0.0524828227807763)
(940,0.0516626681721872)
(960,0.0508779111481614)
(980,0.0508144053858036)
(1000,0.050031323579251)

};
\addlegendentry{$\gamma=0.99$}
\addplot[green!20!blue,sharp plot,mark=*, mark size=0.5pt,mark options={solid}]               
coordinates
{

(0,0.1)
(20,0.19130600963572)
(40,0.164516037056074)
(60,0.221910295315829)
(80,0.0822195313029574)
(100,0.0880946998778492)
(120,0.094729604093957)
(140,0.0847092238100441)
(160,0.0880255979399002)
(180,0.0885235324467827)
(200,0.0898753842469678)
(220,0.103989808978423)
(240,0.106072646991462)
(260,0.119338491765022)
(280,0.137157275992429)
(300,0.142084474485687)
(320,0.152097079937467)
(340,0.158037516330217)
(360,0.179416259322458)
(380,0.192008152624347)
(400,0.169004432961314)
(420,0.177714183740317)
(440,0.169849906841126)
(460,0.174635692132808)
(480,0.161706887617331)
(500,0.15364071314984)
(520,0.149969884192433)
(540,0.149768531210467)
(560,0.136586998421719)
(580,0.110344263082749)
(600,0.105288337646804)
(620,0.0987317475339152)
(640,0.0967922547025228)
(660,0.0915359874318193)
(680,0.0889017090877309)
(700,0.0837911290289768)
(720,0.0824103869844695)
(740,0.0801277774283374)
(760,0.0773451255114839)
(780,0.0732232427882904)
(800,0.0701953794631434)
(820,0.0697679257592644)
(840,0.0679437078046276)
(860,0.0650367899338603)
(880,0.0633403743261401)
(900,0.0621985600184186)
(920,0.0612035279942077)
(940,0.0596905683671497)
(960,0.0584502202230888)
(980,0.0584080979128112)
(1000,0.0572536709164864)

};
\addlegendentry{$\gamma=0.997$}
\end{axis}                         
\end{tikzpicture}

%% file: fig41.tex
\begin{tikzpicture}                 
\begin{axis}[width=8.5cm, height=5.7cm, ymin=0,
ymax=9, tick align=outside, grid=both,
yticklabel style={
/pgf/number format/.cd,
precision=2,
/tikz/.cd
},grid style={line width=.1pt, draw=gray!10},
legend style={nodes={scale=0.55, transform shape}}]                        

\addplot[red!30!green,sharp plot,mark=*, mark size=0.5pt,mark options={solid}]               
coordinates
{

(0,1.35741583589838)
(20,1.35933648650139)
(40,1.35276975232914)
(60,1.34746661798915)
(80,1.34501665111222)
(100,1.33952614598968)
(120,1.3287853069084)
(140,1.30784462273125)
(160,1.30598244235337)
(180,1.3086725996896)
(200,1.3116190331996)
(220,1.30427740163851)
(240,1.29518312283568)
(260,1.29212079026898)
(280,1.29155107261844)
(300,1.29440528208682)
(320,1.29814559984838)
(340,1.29588506682624)
(360,1.287158833377)
(380,1.28139629680913)
(400,1.27963388008176)
(420,1.26492306387037)
(440,1.2631492242939)
(460,1.26039698354972)
(480,1.26824028242021)
(500,1.26743823645019)
(520,1.26531143736546)
(540,1.26682573486964)
(560,1.27390874043793)
(580,1.28028414966214)
(600,1.27822853821749)
(620,1.27685456556913)
(640,1.27312498881956)
(660,1.27439129881576)
(680,1.27115465953791)
(700,1.2680132193754)
(720,1.26998258580479)
(740,1.27029032948045)
(760,1.25905451980172)
(780,1.25479694513266)
(800,1.25079096995679)
(820,1.24457756483616)
(840,1.25033482737406)
(860,1.25005634812859)
(880,1.24919085913426)
(900,1.25233291991354)
(920,1.25322524403789)
(940,1.24958354414791)
(960,1.24440626371625)
(980,1.24178469331295)
(1000,1.23768292743064)

};
\addlegendentry{$\gamma=0.008$}

\addplot[red!20!green,sharp plot,mark=*, mark size=0.5pt,mark options={solid}]               
coordinates
{

(0,1.35741583589838)
(20,1.36312455868496)
(40,1.35103703758476)
(60,1.34138354626678)
(80,1.33806554288964)
(100,1.32797751979287)
(120,1.31023711819874)
(140,1.27157997207232)
(160,1.26990921438549)
(180,1.27578586947148)
(200,1.28165861917037)
(220,1.26995067122405)
(240,1.25330477704329)
(260,1.25156616816506)
(280,1.25267526594293)
(300,1.25870492988399)
(320,1.26604743452172)
(340,1.26125889991215)
(360,1.24448206171123)
(380,1.23553383343358)
(400,1.23211161621461)
(420,1.20368360116911)
(440,1.2020329637659)
(460,1.19869469128405)
(480,1.21521270493797)
(500,1.21484417485893)
(520,1.21105856072895)
(540,1.21487328265274)
(560,1.22685809286582)
(580,1.23853946499664)
(600,1.23661137636313)
(620,1.234079906737)
(640,1.2272932553459)
(660,1.22836079602595)
(680,1.22328776567928)
(700,1.2194841635997)
(720,1.22312131793356)
(740,1.22240112661203)
(760,1.20345294203638)
(780,1.19462487835219)
(800,1.18679498373871)
(820,1.17747767665781)
(840,1.18744421291332)
(860,1.18675751402302)
(880,1.18751347426208)
(900,1.19304354184826)
(920,1.1931241133662)
(940,1.18595349805666)
(960,1.1793282714244)
(980,1.17637280186253)
(1000,1.16919680914317)

};
\addlegendentry{$\gamma=0.1$}
\addplot[red!10!green,sharp plot,mark=*, mark size=0.5pt,mark options={solid}]               
coordinates
{

(0,1.35741583589838)
(20,1.36871368993842)
(40,1.35224980931295)
(60,1.33893258577543)
(80,1.33618282100349)
(100,1.32197357168613)
(120,1.30017076027196)
(140,1.24648894519618)
(160,1.24688397461033)
(180,1.25647792647156)
(200,1.26505432852716)
(220,1.25172428303074)
(240,1.22860060694878)
(260,1.23165963521026)
(280,1.23622956802384)
(300,1.24535357776946)
(320,1.25628370624807)
(340,1.24834506073309)
(360,1.22303789083554)
(380,1.21224632445325)
(400,1.20630822609816)
(420,1.16422748294525)
(440,1.16295275995188)
(460,1.15999287280005)
(480,1.18628740097642)
(500,1.18694756326485)
(520,1.18142346656692)
(540,1.18791842119708)
(560,1.20297213522864)
(580,1.21930995655944)
(600,1.21847896646232)
(620,1.21475745519789)
(640,1.20454649402751)
(660,1.20492858423584)
(680,1.19840750219392)
(700,1.19446673122362)
(720,1.19959699820016)
(740,1.19669519823559)
(760,1.17209449351589)
(780,1.15816598818016)
(800,1.14655159169811)
(820,1.13550517595713)
(840,1.14823016979402)
(860,1.14607754902567)
(880,1.14936939733466)
(900,1.156955915994)
(920,1.15517442348524)
(940,1.1441674271034)
(960,1.13726783809903)
(980,1.13511682758352)
(1000,1.12541337072622)

};
\addlegendentry{$\gamma=0.2$}
\addplot[green!100!blue,sharp plot,mark=*, mark size=0.5pt,mark options={solid}]               
coordinates
{

(0,1.35741583589838)
(20,1.37604915976506)
(40,1.35648561990094)
(60,1.34007245253354)
(80,1.3393066659089)
(100,1.32112224681518)
(120,1.297728833737)
(140,1.2315432545093)
(160,1.23598112515415)
(180,1.24998020773171)
(200,1.26104209815967)
(220,1.24903375149531)
(240,1.22038681073086)
(260,1.23124971808481)
(280,1.24096835223468)
(300,1.25308078587883)
(320,1.26778288800677)
(340,1.25587315842672)
(360,1.22078435718244)
(380,1.20892208571892)
(400,1.19901178150279)
(420,1.14268719095818)
(440,1.14112576363397)
(460,1.13891210114097)
(480,1.17658154821075)
(500,1.1787911419942)
(520,1.17117534422234)
(540,1.18075007284803)
(560,1.19731079342237)
(580,1.21804934528177)
(600,1.21889733955998)
(620,1.21411011872149)
(640,1.19964911623225)
(660,1.19942024280956)
(680,1.19135276813509)
(700,1.18688047175769)
(720,1.19337186159414)
(740,1.1870746410377)
(760,1.15816630465962)
(780,1.13841194229778)
(800,1.12289092855658)
(820,1.11087783451387)
(840,1.12511338815544)
(860,1.12026052456499)
(880,1.12650929640315)
(900,1.13611300190028)
(920,1.13152161327922)
(940,1.11633290448588)
(960,1.10939820543611)
(980,1.10864120429692)
(1000,1.0966350238416)

};
\addlegendentry{$\gamma=0.3$}
\addplot[green!90!blue,sharp plot,mark=*, mark size=0.5pt,mark options={solid}]               
coordinates
{

(0,1.35741583589838)
(20,1.38508807277081)
(40,1.36387043369744)
(60,1.34494234837092)
(80,1.34768631088921)
(100,1.3253771040076)
(120,1.3026752622723)
(140,1.22667121648646)
(160,1.23752626607775)
(180,1.25687986715011)
(200,1.27035602138758)
(220,1.2632509952604)
(240,1.23020442007194)
(260,1.25175689185182)
(280,1.26847235503447)
(300,1.28379041737434)
(320,1.30268866119774)
(340,1.28586732860389)
(360,1.23905397773375)
(380,1.22669550063167)
(400,1.21071309110152)
(420,1.13895034848895)
(440,1.13570721253367)
(460,1.13430402722417)
(480,1.18564624520219)
(500,1.1900071420607)
(520,1.17961853675984)
(540,1.19282430970967)
(560,1.20949627958016)
(580,1.23466723995044)
(600,1.23761813021005)
(620,1.23201792637921)
(640,1.21201577120564)
(660,1.21159299364533)
(680,1.2015138234079)
(700,1.19550099185929)
(720,1.20313165896729)
(740,1.19194065156431)
(760,1.15964162413712)
(780,1.13309848199366)
(800,1.11333828897364)
(820,1.10080503261481)
(840,1.1154698326209)
(860,1.10657015072642)
(880,1.1159738354708)
(900,1.1276818657523)
(920,1.11914234889468)
(940,1.09943279471331)
(960,1.0922318562561)
(980,1.0930979992928)
(1000,1.0787308525298)

};
\addlegendentry{$\gamma=0.4$}
\addplot[green!80!blue,sharp plot,mark=*, mark size=0.5pt,mark options={solid}]               
coordinates
{

(0,1.35741583589838)
(20,1.39579979146663)
(40,1.37458480311321)
(60,1.35387321591808)
(80,1.36192048732815)
(100,1.33504652081635)
(120,1.31541950691319)
(140,1.2328648953141)
(160,1.25332555553813)
(180,1.2793698462247)
(200,1.29553172486297)
(220,1.29815936400191)
(240,1.26242497806898)
(260,1.2978477350826)
(280,1.3238327599804)
(300,1.34345146883118)
(320,1.36738615667838)
(340,1.34472087823832)
(360,1.28349503473955)
(380,1.27128564508854)
(400,1.24607349758696)
(420,1.15687980096923)
(440,1.14971924633176)
(460,1.14910969627488)
(480,1.21742845075548)
(500,1.22463291016942)
(520,1.21025362617119)
(540,1.22790519352261)
(560,1.24338992880893)
(580,1.2731121435257)
(600,1.278468140454)
(620,1.2722081298208)
(640,1.24463744381647)
(660,1.24467058267509)
(680,1.23168230755448)
(700,1.22252247428688)
(720,1.2308378237857)
(740,1.21271214039196)
(760,1.17763606337533)
(780,1.14296920717832)
(800,1.11833054554654)
(820,1.10549477809877)
(840,1.11966331795593)
(860,1.10512458074452)
(880,1.11769223341201)
(900,1.13153926022244)
(920,1.1174614845169)
(940,1.09287562764102)
(960,1.08483711461699)
(980,1.08721732971634)
(1000,1.07013517280441)

};
\addlegendentry{$\gamma=0.5$}
\addplot[green!70!blue,sharp plot,mark=*, mark size=0.5pt,mark options={solid}]               
coordinates
{

(0,1.35741583589838)
(20,1.4081666646183)
(40,1.38887399195927)
(60,1.36742497859958)
(80,1.38305078797407)
(100,1.35087845265539)
(120,1.33721207507362)
(140,1.25263329299095)
(160,1.28744152526155)
(180,1.32208466472978)
(200,1.34197555580342)
(220,1.36147533946968)
(240,1.32626031869516)
(260,1.37981419982351)
(280,1.41821891033358)
(300,1.4453405263006)
(320,1.47632672854521)
(340,1.44735001149741)
(360,1.36830443571829)
(380,1.35722617620285)
(400,1.31735529545184)
(420,1.20714748615852)
(440,1.19260997244651)
(460,1.19313413084047)
(480,1.28324021267005)
(500,1.29396961676113)
(520,1.27350702274081)
(540,1.29694743208496)
(560,1.31007868565099)
(580,1.34395185453303)
(600,1.35184630236961)
(620,1.34431741208168)
(640,1.30571031256286)
(660,1.30699950026159)
(680,1.28945893969881)
(700,1.27464229226967)
(720,1.28271489730552)
(740,1.25467726558844)
(760,1.21712811349503)
(780,1.17224704649408)
(800,1.14154913078445)
(820,1.12837874521081)
(840,1.14133649640633)
(860,1.11898918697876)
(880,1.13440060297383)
(900,1.1500380141433)
(920,1.12799009910541)
(940,1.09809381366666)
(960,1.08826388680713)
(980,1.09157129804389)
(1000,1.07104278973322)

};
\addlegendentry{$\gamma=0.6$}
\addplot[green!60!blue,sharp plot,mark=*, mark size=0.5pt,mark options={solid}]               
coordinates
{

(0,1.35741583589838)
(20,1.42218511715171)
(40,1.40706340148215)
(60,1.38645892426077)
(80,1.41273679438147)
(100,1.37426106874131)
(120,1.37063254154551)
(140,1.29114399919599)
(160,1.34816502438798)
(180,1.39419779912481)
(200,1.42081028020356)
(220,1.46887930448453)
(240,1.44138020829491)
(260,1.52054786165139)
(280,1.5767382596345)
(300,1.6200538948796)
(320,1.66465795616171)
(340,1.63151169253625)
(360,1.53193055312907)
(380,1.52366908968254)
(400,1.45742981480757)
(420,1.31795558397618)
(440,1.29060275186574)
(460,1.29478415183704)
(480,1.41363927936726)
(500,1.42834024452237)
(520,1.39850592983061)
(540,1.43054283542804)
(560,1.44064413145343)
(580,1.4751285145462)
(600,1.48565055984302)
(620,1.47309257381388)
(640,1.41639880489909)
(660,1.4195665437186)
(680,1.39408880216407)
(700,1.36869346055991)
(720,1.37445346534114)
(740,1.33182132363315)
(760,1.29144506124894)
(780,1.23222760991619)
(800,1.193008373914)
(820,1.17900248872617)
(840,1.19042669259923)
(860,1.15664398387684)
(880,1.17372371699604)
(900,1.18949483472079)
(920,1.15539432659338)
(940,1.11953122494903)
(960,1.10630825264761)
(980,1.10901499493668)
(1000,1.08371209764214)

};
\addlegendentry{$\gamma=0.75$}
\addplot[green!50!blue,sharp plot,mark=*, mark size=0.5pt,mark options={solid}]               
coordinates
{

(0,1.35741583589838)
(20,1.43786719530707)
(40,1.42958157557584)
(60,1.41226484220419)
(80,1.45357119072093)
(100,1.40764497651212)
(120,1.42071683042397)
(140,1.35912021983323)
(160,1.45335865981704)
(180,1.51503827635964)
(200,1.55734996543194)
(220,1.65634180514504)
(240,1.65644729762149)
(260,1.78039225756909)
(280,1.86922091011549)
(300,1.95488398521218)
(320,2.04381871326697)
(340,2.02596718618947)
(360,1.92146632428699)
(380,1.91930467755616)
(400,1.78979254854197)
(420,1.59220411374781)
(440,1.54391577105646)
(460,1.56947993295465)
(480,1.72359427724821)
(500,1.74422021208169)
(520,1.70204417336999)
(540,1.75205126887063)
(560,1.76059525751379)
(580,1.77139295854211)
(600,1.7883524857409)
(620,1.74977146349412)
(640,1.65398476120936)
(660,1.65599267405774)
(680,1.61332660003996)
(700,1.55936031424354)
(720,1.55583876734695)
(740,1.48968389642982)
(760,1.44287308893193)
(780,1.35713397286341)
(800,1.30217274656603)
(820,1.28504362035866)
(840,1.29529543065178)
(860,1.24166549803697)
(880,1.25631488462841)
(900,1.26583207921101)
(920,1.21172747425174)
(940,1.16861405773508)
(960,1.14884159510654)
(980,1.14700807347884)
(1000,1.11438883577103)

};
\addlegendentry{$\gamma=0.8$}
\addplot[green!40!blue,sharp plot,mark=*, mark size=0.5pt,mark options={solid}]               
coordinates
{

(0,1.35741583589838)
(20,1.44634063994772)
(40,1.44263420512537)
(60,1.42827889608827)
(80,1.47936825204785)
(100,1.42934988995091)
(120,1.45486987879014)
(140,1.41082844586391)
(160,1.53447512571582)
(180,1.60635404369011)
(200,1.66578023341031)
(220,1.80721371831711)
(240,1.84138776369426)
(260,2.00939885272644)
(280,2.13645904698337)
(300,2.27077104097345)
(320,2.43534570343244)
(340,2.47121960272257)
(360,2.43342276405466)
(380,2.43703884389689)
(400,2.22623632971573)
(420,1.95945068866905)
(440,1.9000409669752)
(460,1.98145593902365)
(480,2.1387275441564)
(500,2.17846975049332)
(520,2.13391994826253)
(540,2.20874727320049)
(560,2.22109471832353)
(580,2.1602820373628)
(600,2.19573398799759)
(620,2.09683856593408)
(640,1.94762605291747)
(660,1.93453952195306)
(680,1.86920849518572)
(700,1.76416938228566)
(720,1.74225287822588)
(740,1.65419684081828)
(760,1.59604080875682)
(780,1.48046502967635)
(800,1.40805083098669)
(820,1.38544010232527)
(840,1.39503047315578)
(860,1.32277314219268)
(880,1.33137582141341)
(900,1.33027782184194)
(920,1.26006146405873)
(940,1.21228979847004)
(960,1.18673611398182)
(980,1.1791089540261)
(1000,1.14100205707507)

};
\addlegendentry{$\gamma=0.9$}
\addplot[green!30!blue,sharp plot,mark=*, mark size=0.5pt,mark options={solid}]               
coordinates
{

(0,1.35741583589838)
(20,1.45469636904008)
(40,1.52609430188872)
(60,1.5355946250525)
(80,1.67770120556028)
(100,1.7637494680079)
(120,1.70489491980346)
(140,1.95641140654553)
(160,1.91005897311857)
(180,1.99366591060112)
(200,2.16420206421521)
(220,1.99816061201615)
(240,2.30771426707011)
(260,2.55365320562131)
(280,2.73513068839604)
(300,2.98693827799842)
(320,3.43979916754767)
(340,4.57417266981134)
(360,7.3048196716652)
(380,10.6483090756326)
(400,17.03691658108048)
(420,23.7892378797375)
(440,26.88782025941229)
(460,40.79773248490334)
(480,50.46250752285425)
(520,50.96038391537381)
(540,40.45480291437645)
(560,24.34358668696953)
(580,15.44590445165684)
(600,8.9869942738906)
(620,8.76778478339427)
(640,5.20355933699284)
(660,4.95173712090443)
(680,3.68246586375582)
(700,3.30149303912706)
(720,3.16688737993478)
(740,3.00692073304719)
(760,2.90154273402852)
(780,2.71010278895866)
(800,2.59497664988951)
(820,1.8540464390689)
(840,1.7596802441801)
(860,1.65331665566743)
(880,1.5468454561553)
(900,1.42264539830874)
(920,1.3291428088304)
(940,1.27552448704355)
(960,1.24112922993276)
(980,1.22312858193)
(1000,1.17769175739615)

};
\addlegendentry{$\gamma=1.00$}

\end{axis}                         
\end{tikzpicture}

%% file: fig6.tex
\begin{tikzpicture}                 
\begin{axis}[width=7.0cm, height=4.0cm, tick align=outside, grid=both,
    grid style={line width=.1pt, draw=gray!10},
    legend style={nodes={scale=0.7, transform shape}},
    ylabel near ticks,
    xlabel near ticks,
    ]                        

\addplot[red,sharp plot,mark=*, mark size=0.5pt,mark options={solid}]               
coordinates
{

(1,0.575)
(2,0.428)
(4,0.367)
(6,0.350)
(8,0.375)
(10,0.346)
(12,0.328)
(14,0.301)
(16,0.295)
(18,0.276)
(20,0.261)
(22,0.248)
(24,0.246)
(26,0.236)
(30,0.231)
(34,0.225)
(38,0.211)
(40,0.206)
(44,0.204)
(48,0.196)
(52,0.203)
(56,0.183)
(60,0.184)
(64,0.188)
};
\addlegendentry{No Dropout}
\addplot[green,sharp plot,mark=*, mark size=0.5pt,mark options={solid}]               
coordinates
{

(1,0.587)
(2,0.442)
(4,0.370)
(6,0.340)
(8,0.297)
(10,0.279)
(12,0.262)
(14,0.255)
(16,0.244)
(18,0.239)
(20,0.233)
(22,0.231)
(24,0.214)
(26,0.207)
(30,0.201)
(34,0.192)
(38,0.185)
(40,0.184)
(44,0.175)
(48,0.168)
(52,0.164)
(56,0.165)
(60,0.152)
(64,0.149)
};
\addlegendentry{With Dropout ($\gamma=0.8$)}
\end{axis}
\end{tikzpicture}

%% file: fig7noise20.tex
\begin{tikzpicture}                 
\begin{axis}[width=7.0cm, height=4.0cm, tick align=outside, grid=both,
    grid style={line width=.1pt, draw=gray!10},
    legend style={nodes={scale=0.6, transform shape}},
    ylabel near ticks,
    xlabel near ticks,
    ]                        

\addplot[red,sharp plot,mark=*, mark size=0.5pt,mark options={solid}]               
coordinates
{

(1,0.643)
(2,0.556)
(4,0.500)
(6,0.536)
(8,0.575)
(10,0.617)
(12,0.609)
(14,0.573)
(16,0.563)
(18,0.538)
(20,0.531)
(22,0.508)
(24,0.498)
(26,0.491)
(30,0.478)
(34,0.462)
(38,0.459)
(42,0.438)
(44,0.434)
(48,0.434)
(52,0.431)
(56,0.416)
(60,0.423)
(64,0.405)
};
\addlegendentry{No Dropout}
\addplot[green,sharp plot,mark=*, mark size=0.5pt,mark options={solid}]               
coordinates
{

(1,0.681)
(2,0.580)
(4,0.510)
(6,0.480)
(8,0.513)
(10,0.524)
(12,0.532)
(14,0.524)
(16,0.523)
(18,0.500)
(20,0.499)
(22,0.473)
(24,0.464)
(26,0.449)
(30,0.430)
(34,0.430)
(38,0.413)
(42,0.406)
(44,0.400)
(48,0.388)
(52,0.392)
(56,0.395)
(60,0.374)
(64,0.374)
};
\addlegendentry{With Dropout ($\gamma=0.8$)}
\addplot[white!40!blue
,sharp plot,mark=*, mark size=0.5pt,mark options={solid}]               
coordinates
{

(1,0.705)
(2,0.591)
(4,0.513)
(6,0.495)
(8,0.478)
(10,0.476)
(12,0.484)
(14,0.484)
(16,0.484)
(18,0.487)
(22,0.471)
(26,0.454)
(30,0.454)
(34,0.437)
(40,0.423)
(48,0.402)
(56,0.408)
(64,0.413)
};
\addlegendentry{With Dropout ($\gamma=0.5$)}
\addplot[white!50!black
,sharp plot,mark=*, mark size=0.5pt,mark options={solid}]               
coordinates
{

(1,0.776)
(2,0.666)
(4,0.568)
(6,0.529)
(8,0.503)
(10,0.508)
(12,0.526)
(14,0.500)
(16,0.496)
(18,0.547)
(22,0.517)
(26,0.499)
(30,0.515)
(34,0.519)
(40,0.502)
(48,0.495)
(56,0.474)
(64,0.456)
};
\addlegendentry{With Dropout ($\gamma=0.2$)}
\end{axis}
\end{tikzpicture}

%% file: fig7traincnn.tex
\begin{tikzpicture}                 
\begin{axis}[width=7.0cm, height=4.0cm, tick align=outside, grid=both,
    grid style={line width=.1pt, draw=gray!10},
    legend style={nodes={scale=0.7, transform shape}}]                        

\addplot[red,sharp plot,mark=*, mark size=0.5pt,mark options={solid}]               
coordinates
{

(1,1.562)
(2,1.106)
(4,0.855)
(6,0.460)
(8,0.255)
(10,0.022)
(12,0.004)
(14,0.000)
(16,0.000)
(18,0.000)
(20,0.000)
(22,0.000)
(26,0.000)
(30,0.000)
(34,0.000)
(38,0.000)
(40,0.000)
(44,0.000)
(48,0.000)
(52,0.000)
(56,0.000)
(60,0.000)
(64,0.000)
};
\addlegendentry{No dropout}
\addplot[green,sharp plot,mark=*, mark size=0.5pt,mark options={solid}]               
coordinates
{

(1,1.616)
(2,1.179)
(4,0.855)
(6,0.460)
(8,0.361)
(10,0.321)
(12,0.206)
(14,0.168)
(16,0.036)
(18,0.018)
(20,0.007)
(22,0.005)
(24,0.001)
(26,0.001)
(28,0.001)
(30,0.000)
(34,0.000)
(38,0.001)
(40,0.001)
(44,0.001)
(48,0.000)
(52,0.000)
(56,0.000)
(60,0.000)
(64,0.000)
};
\addlegendentry{Dropout $\gamma=0.8$}
\end{axis}
\end{tikzpicture}

%% file: fig8train20.tex
\begin{tikzpicture}                 
\begin{axis}[width=7.0cm, height=4.0cm, tick align=outside, grid=both,
    grid style={line width=.1pt, draw=gray!10},
    legend style={nodes={scale=0.7, transform shape}}]                        

\addplot[red,sharp plot,mark=*, mark size=0.5pt,mark options={solid}]               
coordinates
{

(1,1.866)
(2,1.658)
(4,1.393)
(6,1.180)
(8,0.925)
(10,0.883)
(12,0.169)
(14,0.027)
(16,0.009)
(18,0.005)
(20,0.003)
(22,0.003)
(24,0.002)
(26,0.002)
(30,0.001)
(34,0.001)
(38,0.001)
(42,0.001)
(44,0.001)
(48,0.001)
(52,0.001)
(56,0.001)
(60,0.001)
(64,0.001)
};
\addlegendentry{No dropout}
\addplot[green,sharp plot,mark=*, mark size=0.5pt,mark options={solid}]               
coordinates
{

(1,1.936)
(2,1.716)
(4,1.418)
(6,1.165)
(8,0.930)
(10,0.633)
(12,0.380)
(14,0.185)
(16,0.100)
(18,0.050)
(20,0.020)
(22,0.009)
(24,0.006)
(26,0.004)
(30,0.001)
(34,0.001)
(38,0.001)
(42,0.001)
(44,0.001)
(48,0.001)
(52,0.000)
(56,0.000)
(60,0.000)
(64,0.000)
};
\addlegendentry{Dropout $\gamma=0.8$}
\addplot[white!50!blue,sharp plot,mark=*, mark size=0.5pt,mark options={solid}]               
coordinates
{

(1,2.014)
(2,1.795)
(4,1.523)
(6,1.330)
(8,1.091)
(10,0.876)
(12,0.652)
(14,0.447)
(16,0.301)
(18,0.168)
(22,0.055)
(26,0.025)
(30,0.010)
(34,0.005)
(40,0.002)
(48,0.001)
(56,0.001)
(64,0.000)
};
\addlegendentry{Dropout $\gamma=0.5$}
\addplot[white!50!black,sharp plot,mark=*, mark size=0.5pt,mark options={solid}]               
coordinates
{

(1,2.137)
(2,1.989)
(4,1.752)
(6,1.592)
(8,1.422)
(10,1.253)
(12,1.105)
(14,0.959)
(16,0.789)
(18,0.628)
(22,0.482)
(26,0.209)
(30,0.126)
(34,0.050)
(40,0.024)
(48,0.005)
(56,0.003)
(64,0.001)
};
\addlegendentry{Dropout $\gamma=0.2$}
\end{axis}
\end{tikzpicture}